\documentclass[journal]{IEEEtran}
\usepackage[T1]{fontenc}
\usepackage{aecompl}
\usepackage{epstopdf}
\usepackage{cite}
\usepackage{amsmath,amssymb,amsfonts}
\usepackage{algorithmic}
\usepackage{algorithm}
\usepackage{graphicx}
\usepackage{textcomp}
\usepackage{xcolor}
\usepackage{float}
\usepackage{graphicx}
\usepackage{picinpar}
\usepackage{babel}
\usepackage{amsmath}
\usepackage{url}
\usepackage[latin1]{inputenc}
\usepackage{colortbl}
\usepackage{soul}
\usepackage{bm}
\usepackage{multirow}
\usepackage{pifont}
\usepackage{color}
\usepackage{alltt}
\usepackage[hidelinks,colorlinks=true, urlcolor=blue, linkcolor=black, citecolor=black]{hyperref}
\usepackage{enumerate}
\usepackage{siunitx}
\usepackage{breakurl}
\usepackage{epstopdf}
\usepackage{pbox}
\usepackage{authblk}
\usepackage{babel}
\usepackage{mathrsfs,balance,subfigure}
\usepackage{tcolorbox}
\usepackage{subfigure}
\usepackage{stfloats}
\usepackage{booktabs}

\begin{document}
\title{Model-based Chance-Constrained Reinforcement Learning via Separated Proportional-Integral Lagrangian}

\author{Baiyu Peng, Jingliang Duan, Jianyu Chen, Shengbo Eben Li, Genjin Xie, Congsheng Zhang, Yang Guan, Yao Mu, Enxin Sun
\thanks{This work is supported by International Science \& Technology Cooperation Program of China under 2019YFE0100200, NSF China with 51575293, and U20A20334. This work is also partially supported by Geek+. (Corresponding author: Shengbo Eben Li).}% <-this % stops a space
\thanks{Baiyu Peng, Jingliang Duan, Shengbo Eben Li, Genjin Xie, Congsheng Zhang, Yang Guan, Yao Mu and Enxin Sun are with the State Key Laboratory of Automotive Safety and Energy, School of Vehicle and Mobility, Tsinghua University, Beijing, China (e-mail: lishbo@tsinghua.edu.cn).}
\thanks{Jianyu Chen is with the Institute for Interdisciplinary Information Sciences, Tsinghua University, Beijing, China and Shanghai Qi Zhi Institute, Shanghai, China.}
}

\maketitle
	
\begin{abstract}
Safety is essential for reinforcement learning (RL) applied in the real world. Adding chance constraints (or probabilistic constraints) is a suitable way to enhance RL safety under uncertainty. Existing chance-constrained RL methods like the penalty methods and the Lagrangian methods either exhibit periodic oscillations or learn an over-conservative or unsafe policy. 
In this paper, we address these shortcomings by proposing a separated proportional-integral Lagrangian (SPIL) algorithm. We first review the constrained policy optimization process from a feedback control perspective, which regards the penalty weight as the control input and the safe probability as the control output. Based on this, the penalty method is formulated as a proportional controller, and the Lagrangian method is formulated as an integral controller. We then unify them and present a proportional-integral Lagrangian method to get both their merits, with an integral separation technique to limit the integral value in a reasonable range. To accelerate training, the gradient of safe probability is computed in a model-based manner. We demonstrate our method can reduce the oscillations and conservatism of RL policy in a car-following simulation. To prove its practicality, we also apply our method to a real-world mobile robot navigation task, where our robot successfully avoids a moving obstacle with highly uncertain or even aggressive behaviors.

% A demonstration of our method and the experiments can be seen in the video: https://youtu.be/xxxxxxxxxxxxxx

\end{abstract}

\begin{IEEEkeywords}
Safe reinforcement learning, constrained control, robot navigation, neural networks.
\end{IEEEkeywords}

\markboth{\tiny This work has been submitted to the IEEE for possible publication. Copyright may be transferred without notice, after which this version may no longer be accessible}%
{}

% \markboth{Journal of \LaTeX\ Class Files,~Vol.~14, No.~8, August~2015}%
% {Shell \MakeLowercase{\textit{et al.}}: Bare Demo of IEEEtran.cls for IEEE Journals}

\section{Introduction}

\IEEEPARstart{R}{ecent} advances in deep reinforcement learning (RL) have demonstrated state-of-the-art performance in a variety of tasks, including video games \cite{vinyals2019grandmaster, Mnih2015HumanlevelCT ,Hessel2018RainbowCI}, autonomous driving \cite{duan2020hierarchical, ren2020improving} and robotics \cite{kurutach2018model,haarnoja2018soft}. However, most RL successes still remain in virtual environments or simulation platforms. For safety-critical real-world tasks, RL is not yet fully mature or ready to serve as an "off-the-shelf" solution. One of the reasons is the lack of safety constraints \cite{dulac2019challenges}. Consequently, how to handle safety constraints has become a popular and essential topic in RL community.

The safety constraints used in safe RL mainly fall into three categories: expected constraints, worst-case constraints, and chance constraints. Especially, the popular Constrained Markov Decision Process (CMDP) framework \cite{altman1999constrained} is a special case of the expected constraints, which constrains the expected cumulative cost to be below a predetermined boundary.  Many well-known safe RL algorithms build on this framework, including Constrained Policy Optimization (CPO) \cite{achiam2017constrained}, Projection-based Constrained Policy Optimization (PCPO) \cite{yang2019projection}, and Reward Constrained Policy Optimization (RCPO) \cite{tessler2018reward}. However, these methods only guarantee constraint satisfaction in expectation, which is inadequate for safety-critical engineering applications. In this case, the probability of the constraint violation is about 50\% (roughly speaking) \cite{petsagkourakis2020chance}. The second type of constraint is the worst-case constraint, which guarantees constraint satisfaction under any uncertain conditions. Nevertheless, the worst-case constraint tends to be overly conservative, and it only supports systems with bounded noise \cite{dulac2019challenges}. The third form is the chance constraint, where the constraint holds with a predefined probability. The chance constraint clearly limits the occurring probability of the unsafe event, which is selected according to the different demands of users and the tasks. Therefore, the chance constraint is quite suitable for various real-world applications. In this paper, we will focus on the chance-constrained RL problems, i.e., how to learn an optimal policy satisfying the chance constraints. 
%Safe Reinforcement Learning Using Robust MPC

Next, we will briefly introduce the existing chance-constrained RL studies. In 2005, Geibel and Wysotzki use an indicator function to estimate the safe probability by sampling, and add a large penalty term in the reward function if the safe probability is low \cite{Geibel2005RiskSensitiveRL}. Then, the reshaped reward is optimized by an actor-critic method. Giuseppi and Pietrabissa (2020) view the reward and safe probability  as two objectives, and propose a corresponding multi-objective RL method \cite{giuseppi2020chance}.  Paternain et al. (2019)  derive a lower bound of the safe probability, which is employed to construct a surrogate constraint since it has an additive structure and easier to tackle \cite{Paternain2019LearningSP}. Then, the transformed problem is solved by the Lagrangian method, which introduces a Lagrangian multiplier to balance policy performance and constraints satisfaction. To reduce the conservatism introduced by constraining a lower bound, Peng et al. (2020) only use the lower bound to obtain an update direction, but still evaluate the feasibility of the policy using original chance constraints\cite{Peng2020ModelBasedAW}. In addition, they employ a penalty method with increasing weight to enforce constraint satisfaction. 

\begin{figure}[hbt]
    \centering
    \subfigure[Penalty method]{
        \includegraphics[width=0.2057\textwidth]{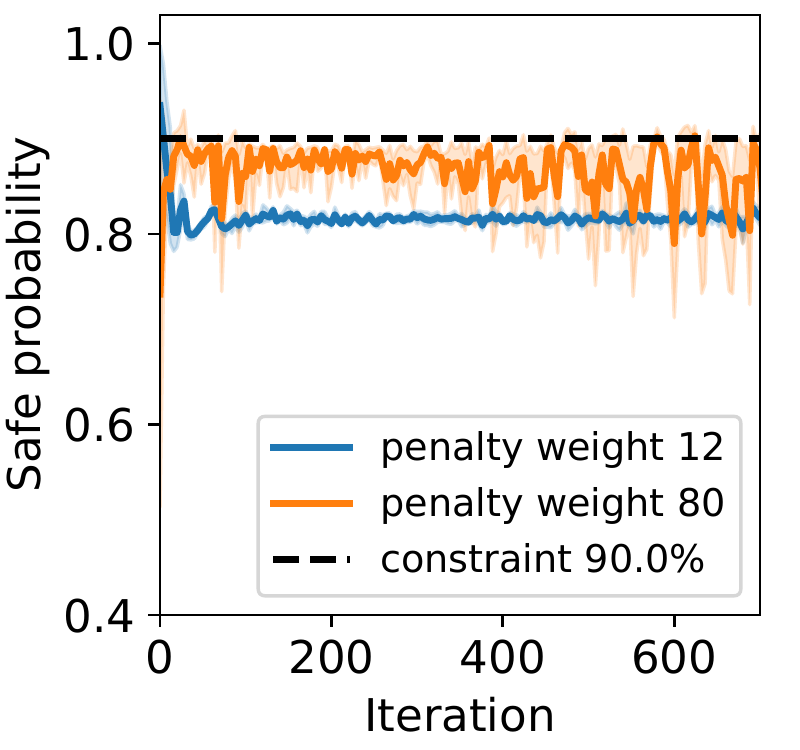}
        \label{demo_penalty0.9}
    }
    \subfigure[Lagrangian method]{
	\includegraphics[width=0.2443\textwidth]{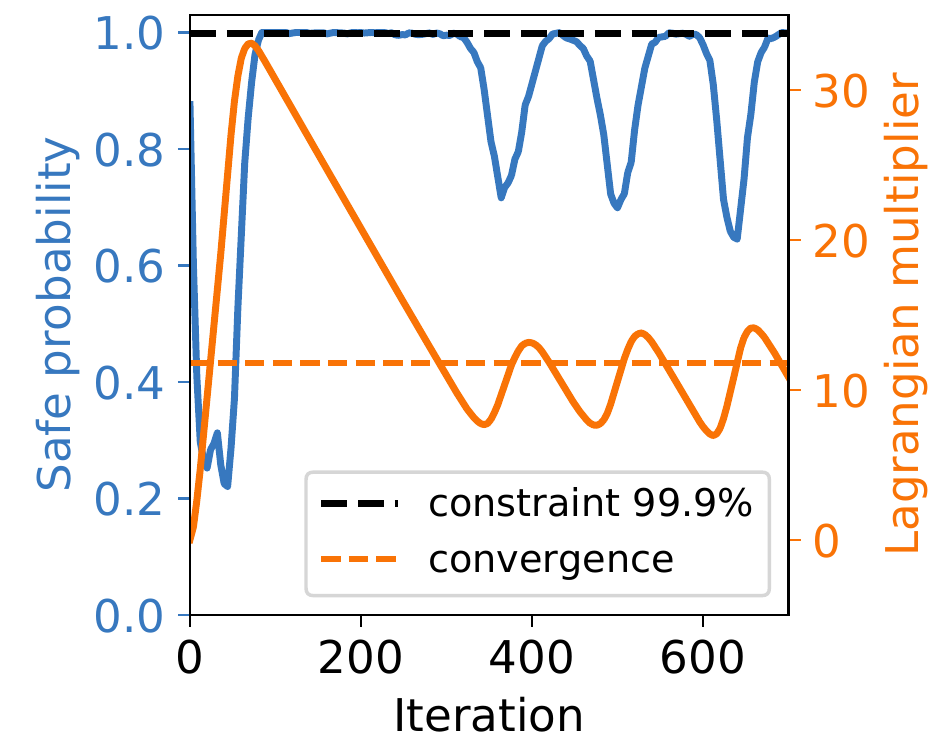}
        \label{demo_Lag0.9}
    }
    \caption { Examples of learning curves for penalty and Lagrangian methods. (a) Penalty method exhibits oscillations and violates the constraint. (b) Lagrangian method exhibits Lagrange multiplier overshooting and oscillations, and further harms the policy learning.}
\end{figure}

The previous chance-constrained RL mainly relies on the penalty method or the Lagrangian method. However, they actually face several challenges such as poor policy performance, constraint violations, and unstable learning  process\cite{farina2016stochastic}. The penalty methods require a well-designed penalty weight to balance the reward and constraint, which unfortunately is non-trivial and hard to tune. As shown in Fig. \ref{demo_penalty0.9}, a large penalty is prone to rapid oscillations and frequent constraint violations, while a small penalty always violates the constraint seriously \cite{tessler2018reward}. 
As for the Lagrangian method, it usually suffers from the Lagrange multiplier overshooting (see Fig. \ref{demo_Lag0.9}), which will lead to an overly conservative policy \cite{Paternain2019LearningSP,chow2017risk}. Besides, due to the delay between the policy optimization and the Lagrange multiplier adaptation, the Lagrangian multiplier usually oscillates periodically during the training process, which further results in policy oscillations \cite{wah2000improving, stooke2020responsive}. In addition, most existing methods do not support model-based optimization since they all rely on a non-differentiable indicator function to estimate the safe probability. Therefore, they can only use model-free methods to optimize the safe probability, which is generally slower than model-based methods.

To overcome the problems mentioned above, we propose a model-based Separated Proportional-Integral Lagrangian (SPIL) method for chance-constrained RL, which can fulfill the safety requirements with a steady and fast learning process. The contributions of this paper are summarized as follows,
\begin{enumerate}
\item 
This paper interprets the chance-constrained policy optimization process from a feedback control perspective, which regards the penalty weight as the control input and the safe probability as the control output. Based on this, we unify the existing constraint optimization methods into the PID control methodology, in which the penalty method is formulated as a proportional controller and the Lagrangian method is formulated as an integral controller. Then, we develop the proportional-integral Lagrangian (PIL) framework by combining the proportional and integral modules to get their both merits.
\item To prevent policy from over-conservatism caused by multiplier overshooting, we draw inspiration from PID control and introduce an integral separation technique, which separates the integrator out when the integral value exceeds a predetermined threshold. Then, we embed this technology into the PIL framework to propose the SPIL method for chance-constrained RL. Simulations demonstrate our SPIL achieves better policy performance compared with existing penalty and Lagrangian methods \cite{Geibel2005RiskSensitiveRL, Paternain2019LearningSP,chow2017risk}. 
\item To accelerate the training process in a model-based way, we adopt an approximated model-based gradient of the safe probability to participate in the policy optimization. The approximated gradient is proved to approach to the true gradient under mild conditions. Compared with a popular model-free method, the learning speed is improved by at least five times.
\item Finally, a real-world mobile robot navigation experiment proves the effectiveness of model-based SPIL algorithm in practical engineering problems.
\end{enumerate}

This article is further organized as follows. The chance-constrained RL problem is formulated in Section \ref{sec:Preliminary}. The model-based SPIL method is proposed in Section \ref{sec:chance constrained algorithm}. In Section \ref{sec:Simulation of Car-following}, our method is verified and compared in a car-following simulation. In Section \ref{sec:Experiment of AGV Obstacle Avoidance}, the mobile robot navigation experiment is presented to prove the practicability of the algorithm. Section \ref{sec:Conclusion} concludes this paper.

\section{Preliminaries}
\label{sec:Preliminary}
\subsection{Problem Description}
In our model-based RL setting, we assume a dynamic model is already available, either learned by collected data or derived by prior knowledge. To indicate the gap between the model and reality, an uncertain term is included in this model. This assumption is reasonable in many physical engineering problems like autonomous driving, where there are many established dynamic models. Therefore, we can directly learn a policy through the given uncertainty model. To ensure the policy feasibility for real environments, a chance constraint needs to be introduced in the model-based learning process. In other words, given a reasonable uncertain term, if the policy is safe with a high probability under model uncertainty, we can usually ensure its safety in practical environments. This setting is similar to that in stochastic control \cite{Mesbah2016StochasticMP}.

The dynamic model and the chance constraint are mathematically described as:
\begin{equation}
\begin{aligned}
&s_{t+1}=f(s_{t}, a_{t}, \xi_{t}), \quad  \xi_{t}\sim p(\xi_{t}),\\
&{\rm Pr}\left\{ \bigcap_{t=1}^{N} \left[h\left(s_{t}\right)<0\right]\right\}\ge1-\delta
\end{aligned}
\end{equation}
where $t$ is the current step, $s_{t}\in{\mathcal{S}}$ is the state, $a_{t}\in{\mathcal{A}}$ is the action, $f(\cdot,\cdot,\cdot)$ is the environmental dynamic, $\xi_{t}\in{\mathbb{R}}^{n}$ is the model uncertainty following an independent and identical distribution $p(\xi_{t})$. $h(\cdot)$ is the safety function defining a safe state region. Here the chance constraint takes a joint form, which is initially brought from stochastic systems control \cite{Mesbah2016StochasticMP}. Intuitively, it requires the probability of being safe over the finite horizon $N$ to be at least $1-\delta$. For simplicity, we only consider one constraint, but our method can readily generalize to multiple constraints.

The chance-constrained problem is defined as maximizing a objective function $J$, i.e., the expected discounted cumulative reward, while keeps a high safe probability $p_s$:
\begin{equation}
\begin{aligned}
\max _{\pi} &J\left(\pi\right)={\mathbb{E}_{s_0,\xi}}\left\{\sum_{t=0}^{\infty} \gamma^{t} r\left(s_{t}, a_{t}\right)\right\}\\
\text { s.t. }&p_s(\pi)={\rm Pr}\left\{ \bigcap_{t=1}^{N} \left[h\left(s_{t}\right)<0\right]\right\}\ge1-\delta, \\
&s_{t+1}=f(s_{t}, a_{t}, \xi_{t}), \quad  \xi_{t}\sim p(\xi_{t})
\label{CCRL problem}
\end{aligned}
\end{equation}
where $r(\cdot,\cdot)$ is the reward function, $\gamma\in(0,1)$ is the discount factor, ${\mathbb{E}_{s_0,\xi}}(\cdot)$ is the expectation w.r.t. the initial state $s_0$ and uncertainty $\xi_{0:\infty}$. $\pi$ is a deterministic policy, i.e., a mapping from state space ${\mathcal{S}}$ to action space ${\mathcal{A}}$. In practice, policy is usually a parameterized neural network with parameters $\theta$, denoted as $\pi(s_{t};\theta)$ or $\pi_{\theta}$. 

\subsection{Penalty and Lagrangian Methods}
{
To find the optimal control policy for problem \eqref{CCRL problem}, the penalty and Lagrangian methods are widely employed in existing studies \cite{Geibel2005RiskSensitiveRL,giuseppi2020chance, Paternain2019LearningSP, Peng2020ModelBasedAW}. The penalty method adds a quadratic penalty term in the objective function to force the satisfaction of the constraint:
\begin{align}
\label{penalty}
    &\max_\pi J(\pi) - \frac{1}{2}K_P \left((1-\delta- p_s(\pi))^+\right)^2
\end{align}
where $K_p>0$ is the penalty weight, $p_s(\pi)$ is the joint safe probability (see \eqref{CCRL problem}), and $(\cdot)^+$ means $\max(\cdot,0)$.
This unconstrained problem is usually solved by gradient ascent:
\begin{equation}
\label{penalty_update}
    \theta^{k} \leftarrow \theta^{k-1}+\alpha_{\theta}(\nabla_{\theta}J^{k-1}+K_P (1-\delta - p_s^{k-1})^+ \nabla_{\theta}p_s^{k-1})
\end{equation}
where $k$ means the $k$-th iteration, $\alpha_{\theta}>0$ is the learning rate, $J^{k}$ and $p_s^{k}$ are short for $J(\pi_{\theta^k})$ and $p_s(\pi_{\theta^k})$ respectively.  For practical applications, it is usually difficult to select an appropriate weight $K_P$ to balance reward and constraint well. A large penalty is prone to rapid oscillations and frequent constraint violations, while a small penalty always violates the constraint seriously.  

For the Lagrangian method, it first transforms the chance-constrained problem \eqref{CCRL problem} into an dual problem by introduction of the Lagrange multiplier $\lambda_I$ \cite{boyd2004convex}:
\begin{align}
\label{maxmax}
    &\max_{\lambda_I\ge0}\min_\pi\mathcal{L}(\pi,\lambda_I)=-J(\pi)+\lambda_I\left(1-\delta- p_s(\pi)\right)
\end{align}
Then, problem \eqref{maxmax} can be
solved by dual ascent, i.e., alternatively update the Lagrange multiplier and primal variables:
\begin{equation}
\label{dual_update}
    \lambda_I^{k}\leftarrow (\lambda_I^{k-1} +K_{I}(1-\delta- p_s^{k-1}))^+,
\end{equation}
\begin{equation}
\label{primal_update}
    \theta^{k} \leftarrow \theta^{k-1}+\alpha_{\theta}(\nabla_{\theta}J^{k-1}+\lambda_I^k \nabla_{\theta}p_s^{k-1})
\end{equation}
where $K_{I}>0$ is the learning rate for $\lambda_I$. 

As mentioned in Introduction, the Lagrangian method usually faces the Lagrange multiplier overshooting and multiplier oscillation challenges, resulting in poor policy performance and unstable learning process.
}

\section{Methodology}
\label{sec:chance constrained algorithm}
In this section, we first reshape the penalty method and Lagrangian method in a feedback control view. Then we unify them to formulate a proportional-integral Lagrangian method to improve the policy performance without losing the safety requirement. Finally, we introduce an integral separation technique and a model-based gradient to make the whole method practical and efficient. 

\subsection{Feedback Control View of Chance-Constrained Policy Optimization}\label{SPIL}

The key insight of the proposed method comes from a deep and novel understanding of the penalty and Lagrangian methods from a control perspective. From \eqref{penalty_update} and \eqref{primal_update}, the update rule of existing methods can be expressed in a similar form
\begin{equation}
\label{unified update}
\theta^{k} \leftarrow\theta^{k-1}+\alpha_{\theta}(\nabla_{\theta}J^{k-1}+\lambda^k \nabla_{\theta}p_s^{k-1})
\end{equation}
where $\lambda ^k$ is actually a balancing weight. The core difference between the two methods lies in the selection of $\lambda$. 
For the penalty method, $\lambda^k=K_P (1-\delta - p_s^{k-1})^+$, which indicates the constraint violation at the $k$-th iteration. For the Lagrangian method, $\lambda^k=\lambda_I^k$ in \eqref{dual_update}, which can be regarded as the sum of constraint violation over previous $k$ iterations. This insight inspires us to review RL from a control perspective.

\begin{figure}[htbp]
\centering 
\centerline{\includegraphics[width=0.45\textwidth]{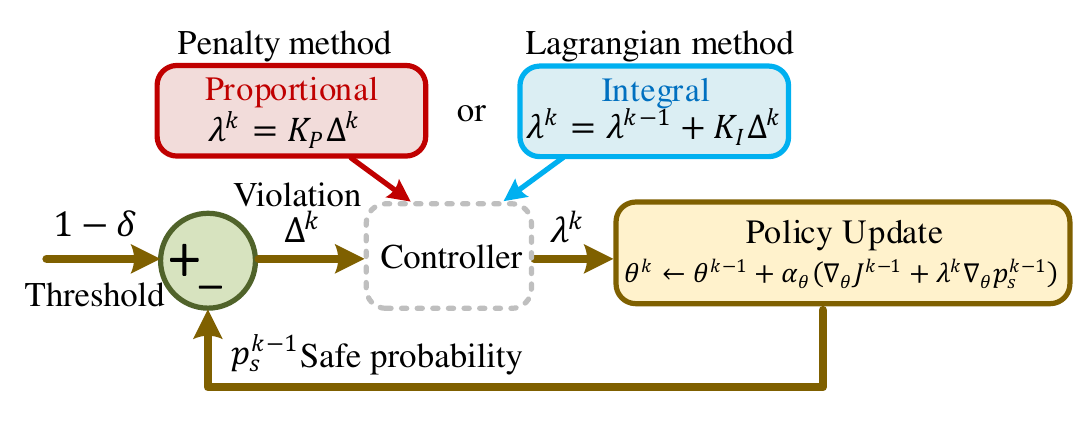}}
\caption{Feedback control view of chance-constrained policy optimization.}
\label{fig:feedback view}
\end{figure}

As shown in Fig. \ref{fig:feedback view}, one can view the policy optimization as a feedback control process, where $\theta^k$ is the state, $\lambda^k$ is the control signal, policy update \eqref{unified update} is the system dynamics (state transition equation), $p_s^k$ is the system output, $1-\delta$ is the desired output and $1-\delta- p_s^{k-1}$ is the tracking error (or constraint violation). Then, the essence of this system is the design of the controller, i.e., given the tracking error $1-\delta- p_s$, how can we decide the control signal $\lambda$? The penalty method actually adopts $\lambda^k$ proportional to the violation. The Lagrangian method instead computes $\lambda^k$ as the sum of previous constraint violations. In such an insight, the penalty method becomes exactly a proportional (P) controller, and the Lagrangian method becomes an integral (I) controller. Subsequently, one can easily understand the merits and flaws of these two methods by analogy. For pure ``P" control, small $K_P$ leads to steady-state error, while large $K_P$ exhibits oscillation. This matches the phenomenons we observe in the penalty method. Similarly, the problems of overshooting and oscillation in Lagrangian method can also be explained by properties of pure ``I" control.

% Comparing the policy update rule of penalty method \eqref{penalty_update} with that of Lagrangian method \eqref{primal_update}, one may find they are surprisingly similar. Both gradients are the weighted sums of  $\nabla_{\theta}J^{k}$ and $\nabla_{\theta}p_s^{k}$. The only difference lies in that the weight $(1-\delta + p_s^k)^+$ in penalty method is the constraint violation at $k$-th iteration, while the weight $\lambda$ in Lagrangian method is the constraint violation accumulated in the previous $k$ iterations. This insight builds the bridge between optimization and feedback control. As shown in Fig. \ref{fig:feedback}, one can view the optimization as feedback control, where  $p_s$ is the system output, $1-\delta$ is the desired output , $1-\delta- p_s^k$ is the tracking error and $\lambda$ is the control signal. Consequently, the penalty method becomes a proportional controller, where it determines the Lagrangian multiplier proportional to constraint violation. While the Lagrangian method becomes an integral controller, where it determines the Lagrangian multiplier as the integral of constraint violation. 

With this insight, we naturally propose to combine the penalty method and Lagrangian method by computing $\lambda^k$ as a weighted sum of proportional and integral values, which leads to the proportional-integral Lagrangian method (PIL). The update process of PIL at $k-$iteration is given as:
\begin{align}
\label{P_update}
    &\Delta^{k} \leftarrow 1-\delta-p_s^{k-1}, \\
\label{I_update}
    &I^{k} \leftarrow (I^{k-1} + \Delta^{k})^+, \\
% \label{D_update}
%     &\Psi^{k} \leftarrow (p_s^{k-1} - p_s^{k})^+ \\
\label{lambda_update}
    &\lambda^{k} \leftarrow (K_P\Delta^k+K_II^k)^+ ,\\
\label{PI_update}
    &\theta^{k} \leftarrow \theta^{k-1}+\frac{\alpha_{\theta}}{1+\lambda^k}(\nabla_{\theta}J^{k-1}+\lambda^k \nabla_{\theta}p_s^{k-1})
\end{align}
where $\Delta^k$ and $I^k$ are proportional and integral values, respectively, with $K_P$ and $K_I$ as coefficients. The proportional term $K_P\Delta^k$ serves as an immediate feedback of the constraint violation. The integral term $K_II^k$ eliminates the steady-state error at convergence. In such a framework, the penalty method and the Lagrangian method can be regarded as two special cases of PIL with $K_P>0, K_I=0$ and $K_P=0,K_I>0$, respectively. Note that, to maintain a relatively consistent step size and make policy update more stable, the gradient for $\theta^k$ in \eqref{PI_update} is re-scaled by $\frac{1}{1+\lambda^k}$.

This mechanism is expected to realize a steady learning process, just like how the proportional-integral controller works. However, there are still two key issues for practical applications:
\begin{enumerate}[1)]
	\item Integral value $I$ easily gets overly large, leading to policy over-conservatism. 
	\item $\nabla_{\theta}p_s^{k}$ is hard to compute, especially in a model-based paradigm.
\end{enumerate}
The following subsections will explain and solve these two problems.

\subsection{Integral Separation Technique}\label{separate}
The integral value $I^k$ increases according to the constraint violation $\Delta^k$. However, when the initial policy is relatively unsafe, $\Delta^k$ can be quite large, which will cause the overshooting of $I^k$ and $\lambda^k$. With a large $\lambda^k$ in \eqref{PI_update}, the policy tends to be overly conservative since the weight of $\nabla_{\theta}p_s^{k-1}$ is overly large. Even worse, since the maximal safe probability is 1, the overshooting and conservatism problems can hardly recover by themselves. For e.g., suppose $1-\delta=0.999, p_s^k=1$, and $\lambda^k$ is already overshooting, the integral term $I^k$ can only fall slowly with the speed of $\Delta^{k}=-0.001$. Therefore, the policy in such a case will deteriorate in a long time. This issue is also not well recognized and resolved in previous similar works like \cite{stooke2020responsive}. 

\begin{figure}[htbp]
\centering 
\centerline{\includegraphics[width=0.45\textwidth]{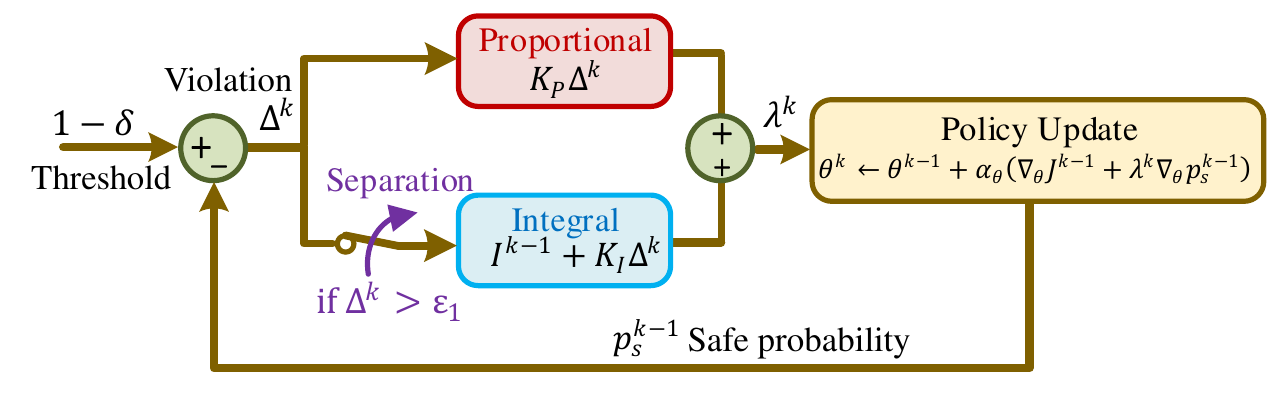}}
\caption{The framework of the proposed SPIL method.}
\label{fig:SPIL algorithm}
\end{figure}

To deal with the overshooting problem of $I^k$ and $\lambda^k$, we draw inspiration from PID control \cite{Jia2019AnIP} and introduce an integral separation technique. As shown in Fig. \ref{fig:SPIL algorithm}, the integrator will be activated only when $\Delta^k$ is less than a certain value. If $\Delta^k$ is too large, the integrator will be blocked to limit the increase of $I^k$. Specifically, \eqref{I_update} is modified to:
\begin{equation}
\begin{aligned}
\label{SI_update}
    &I^{k} \leftarrow (I^{k-1} + K_S\Delta^{k})^+, \\
    &K_S=
    \begin{cases}
    0& \text{$\varepsilon_1<\Delta^{k}$} \\
    \beta& \text{$\varepsilon_2<\Delta^{k}\leq\varepsilon_1$} \\
    1& \text{$\Delta^{k}\leq\varepsilon_2$}
    \end{cases}
\end{aligned}
\end{equation}
where $K_S$ is the separation function, $1>\beta>0$ and $\varepsilon_1>\varepsilon_2>0$ are predetermined parameters. The piece-wise function $K_S$ separates the integrator out or slows it down if the error is relatively large. Such a recipe restrains the occurrence of overshooting and over-conservatism. Our simulation indicates it greatly improves the performance for a large safety threshold $1-\delta$, such as $1-\delta=99.9\%$. We refer to the combination of PIL and the separation technique as the separated proportional-integral Lagrangian (SPIL) method.

\subsection{Model-based Gradient of Safe Probability}
\label{model-based gradient}
Next, we will figure out how to estimate $\nabla_\theta p_s$ in \eqref{PI_update} in an efficient way. Generally, there is not any analytic solution for a joint probability  and its gradient \cite{Mesbah2016StochasticMP, Geletu2019AnalyticAA}, i.e., they are nearly intractable. Therefore, previous researchers usually introduce an indicator function to estimate the probability through sampling  \cite{Paternain2019LearningSP, Peng2020ModelBasedAW}. Due to the discontinuity of the indicator function, these methods are mostly model-free, which are generally believed to be slower than their model-based counterparts \cite{deisenroth2011pilco,Shengbo2019}. 

Inspired by recent advances in stochastic optimization  \cite{Geletu2019AnalyticAA}, we introduce a model-based alternative to $\nabla_{\theta}p_s$, which enables us to estimate $\nabla_{\theta}p_s$ efficiently. We first define an indicator-like function $\phi(z,\tau)$:
\begin{equation}
\begin{aligned}
    &\phi(z,\tau)=\frac{1+b_1\tau}{1+b_2\tau\exp(-\frac{z}{\tau})}, \\
    &0<b_2<\frac{b_1}{1+b_1},  0<\tau<1
\end{aligned}
\end{equation}
where $z$ and $\tau$ are scalar variables of the function, $b_1,b_2$ are the parameters. The expected production of $\phi(z,\tau)$ over $N$ horizon is defined as:
\begin{equation}
    \Phi(\pi,\tau)={\mathbb{E}_{s_0,\xi}}\left\{\prod_{t=1}^{N}\phi\left(-h(s_{t}),\tau\right)\right\}
    \label{expected production}
\end{equation}

\begin{figure}[htbp]
\centerline{\includegraphics[width=0.3\textwidth]{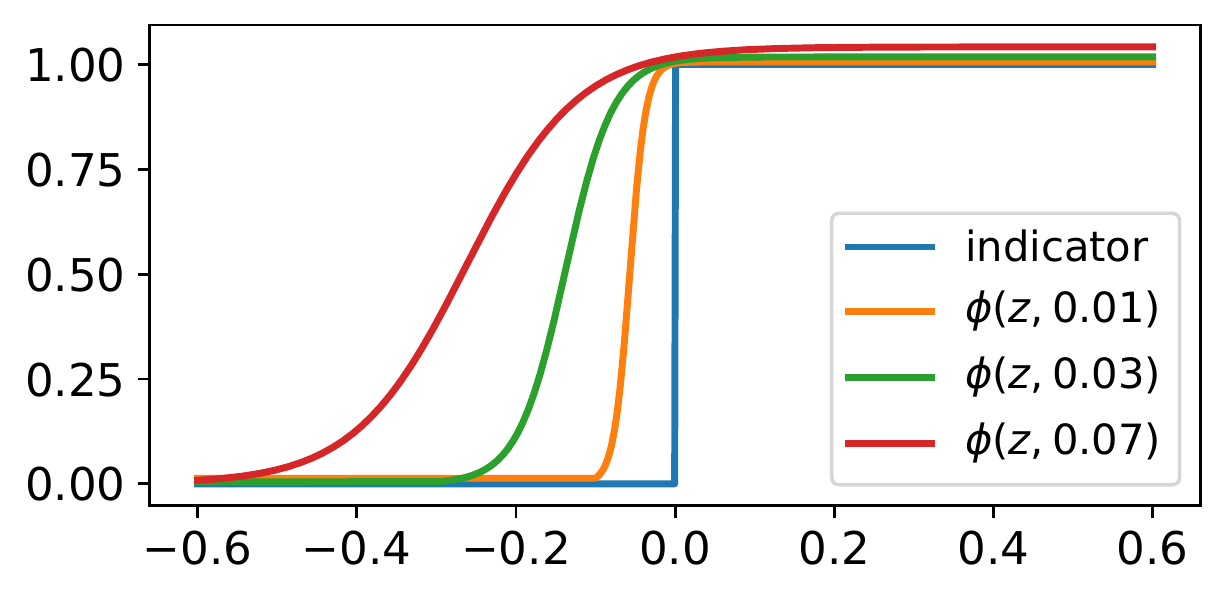}}
\caption{Comparison of indicator function and $\phi(z, \tau)$ with different $\tau$. }
\label{fig:indicator}
\end{figure}

As shown in Fig. \ref{fig:indicator}, $\phi(z,\tau)$ can be intuitively regarded as a differentiable approximation of indicator function for constraint violation, and its expected product $\Phi(\pi,\tau)$ approximates joint safe probability. (To see this, one can image $\phi(z,\tau)$ in \eqref{expected production} as the indicator function. Then its expected production is actually the joint safe probability.)  The parameter $\tau$ controls how well the indicator function is approximated.
Intriguingly, this approximation is not only an intuitive trick and it does have strong theoretical support. Regardless of the nonlinearity and nonconvexity of $h(s_t)$, the gradient of $\Phi(\pi,\tau)$ is proved to converge to the true gradient of joint safe probability $p_s(\pi)$ as $\tau$ approaches $0$ under mild assumptions \cite{Geletu2019AnalyticAA}:
\begin{equation}
\label{app_gradient}
   \lim\limits_{\tau\to0+}
\sup\limits_{\theta\in\Theta}\nabla_{\theta}\Phi(\pi,\tau)=\nabla_{\theta}p_s(\pi)
\end{equation}
where $\Theta$ is a small ball around the current policy network parameter.  For simplicity, we omit mathematical details; interested readers are recommended to refer to \cite{Geletu2019AnalyticAA} for a rigorous explanation. 

In practice, one only needs to pick a relatively small fixed $\tau$ and compute $\nabla_{\theta}\Phi(\pi,\tau)$ to substitute $\nabla_{\theta}p_s(\pi)$, where the expectation is estimated by sampling average. Therefore, the original policy update rule of SPIL is rewritten as:
\begin{equation}
\label{eq.MB-policy update}
\theta^{k} \leftarrow \theta^{k-1}+\frac{\alpha_{\theta}}{1+\lambda^k}(\nabla_{\theta}J^{k-1}+\lambda^k \nabla_{\theta}\Phi(\pi_{\theta^{k-1}},\tau))
\end{equation}
% Obviously, a overly small $\tau$ will harms the computation of  $\nabla_{\theta}\Phi(\pi,\tau)$. Finally, the order of magnitude of $J$ and $p_s$ are usually quite different, which may damage the optimization process. An optional solution is to rescale the gradient $\nabla_{\theta}p_s$ to match the scale of $\nabla_{\theta}J$: 
% \begin{equation}
%     \nabla_{\theta}p_s \leftarrow \frac{\|\nabla_{\theta}J\|}{\|\nabla_{\theta}p_s\|}\nabla_{\theta}p_s
% \end{equation}

It should be pointed out that the introduction of $\phi(z,\tau)$ in this subsection is only used for the computation of $\nabla_{\theta}p_s(\pi)$. The safe probability $p_s$ itself is still estimated through Monte-Carlo sampling, i.e., suppose there are $m$ safe trajectories among all the $M$ trajectories, then the safety probability is $p_s\approx \frac{m}{M}$.

\subsection{Model-based SPIL Algorithm}
\label{actor-critic}
Based on aforementioned gradient, we will propose the model-based SPIL algorithm for practical applications. Firstly, we define the state-action value of $(s,a)$ under policy $\pi$ as:
\begin{align}
    &Q^{\pi}(s,a)={\mathbb{E}_{\xi,\pi}}\left\{\sum_{t=0}^{\infty} \gamma^{t} r\left(s_{t}, a_{t}\right)\Big|s_0=s,a_0=a\right\}
\end{align}
Thus, the expected cumulative reward $J$ in \eqref{CCRL problem} can be expressed as a $N$-step form:
\begin{align}
\label{n-step}
    &J(\pi)={\mathbb{E}_{s_0,\xi}}\left\{\sum_{t=0}^{N-1}\gamma^{t} r\left(s_{t}, a_{t}\right)+\gamma^{N}Q^{\pi}(s_N,s_N)\right\}
\end{align}

For large and continuous state spaces, both value function and policy are parameterized:
\begin{equation}
Q(s,a) \cong Q(s,a ; w), \quad a \cong \pi(s ;\theta)
     \label{para}
\end{equation}
The parameterized state-action value function with parameter $w$ is usually named the ``critic'', and the parameterized policy with parameter $\theta$ is named the ``actor'' \cite{Shengbo2019}.

The parameterized critic is trained by minimizing the average square error:
\begin{equation}
\begin{aligned}
  J_{Q}^k={\mathbb{E}}_{s_0,\xi}\left\{\frac{1}{2}\left(Q_{\text {target}}-Q(s_0,a_0;w^k)\right)^{2}\right\}
     \end{aligned}
\label{td error}
\end{equation}
where $Q_{\text {target}} = \sum_{t=0}^{N-1} \gamma^{t} r\left(s_{t}, a_{t}\right)+\gamma^{N} Q\left(s_{N},a_{N};w^k\right)$ is the $N$-step target. The rollout length $N$ is identical to the horizon of chance constraint.

% The semi-gradient of the critic is
%  \begin{equation}
% \begin{aligned}
%   \nabla_{\omega}J_Q^k&={\mathbb{E}}_{s_0,\xi}\left\{\left(Q(s_0,a_0;w^k)-Q_{\text {target}}\right) \frac{\partial Q(s_0,a_0;w^k)}{\partial w}\right\}
% \end{aligned}
% \label{semi-gradient of the critic}
% \end{equation}

The parameterized actor is trained via gradient ascent in \eqref{eq.MB-policy update}. In particular, we first compute $J$ and $\Phi$ through model rollout. Then $\nabla_{\theta}J$ and $\nabla_{\theta}p_s$ are computed via backpropagation though time with the dynamic model \cite{Shengbo2019}. In practice, this process can be easily finished by any autograd package. The pseudo-code of the proposed algorithm is summarized in Algorithm \ref{alg:SPIL}. 

\begin{algorithm}[!htb]
{
\caption{SPIL algorithm}
\label{alg:SPIL}
\begin{algorithmic}
\STATE Initialize $\pi_{\theta^0}$, $Q_{w^0}$, $I^0=0$, $s_{0}\in \mathcal{S}$, $k=1$ 
\REPEAT
\STATE{Rollout $M$ trajectories by $N$ steps using policy $\pi^{k-1}$}
\STATE{Estimate safe probability}
\STATE{\quad $p_s^{k-1}\leftarrow\frac{m}{M}$}
% \STATE{\quad $J^k=\frac{1}{M}{\sum^{M}}\left\{\sum_{t=0}^{N-1}\gamma^{t} r\left(x_{t}, u_{t}\right)+\gamma^{N}Q^{\pi}(x_N,u_N)\right\}$}
\STATE{Update $\lambda$ via SPIL rules}
\STATE{\quad  $\Delta^{k} \leftarrow 1-\delta-p_s^{k-1}$}
\STATE{\quad  $I^{k} \leftarrow (I^{k-1} + K_S\Delta^{k})^+$}
% \STATE{\quad  $\Psi^{k} \leftarrow (p_s^{k-1} - p_s^{k})^+$}
\STATE{\quad  $\lambda^{k} \leftarrow (K_P\Delta^k+K_II^k)^+ $}
\STATE {Update critic:}
\STATE{\quad  $\omega^{k} \leftarrow \omega^{k-1} + \alpha_{\omega}\nabla_{\omega}J_{Q}^{k-1}$ }
\STATE {Update actor:}
\STATE{\quad  $\theta^{k} \leftarrow \theta^{k-1}+\frac{\alpha_{\theta}}{1+\lambda^k}\left(\nabla_{\theta}J^{k-1}+\lambda^{k} \nabla_{\theta}\Phi(\pi_{\theta^{k-1}},\tau)\right)$ }

\STATE $k\leftarrow k+1$
\UNTIL $|Q^{k}-Q^{k-1}|\le \zeta$ and $|\pi^{k}-\pi^{k-1}| \le \zeta$
\end{algorithmic}}
\end{algorithm}

\section{Simulation Validation}

\label{sec:Simulation of Car-following}
\subsection{Example Description}
\begin{figure}[htbp]
\centerline{\includegraphics[width=0.35\textwidth]{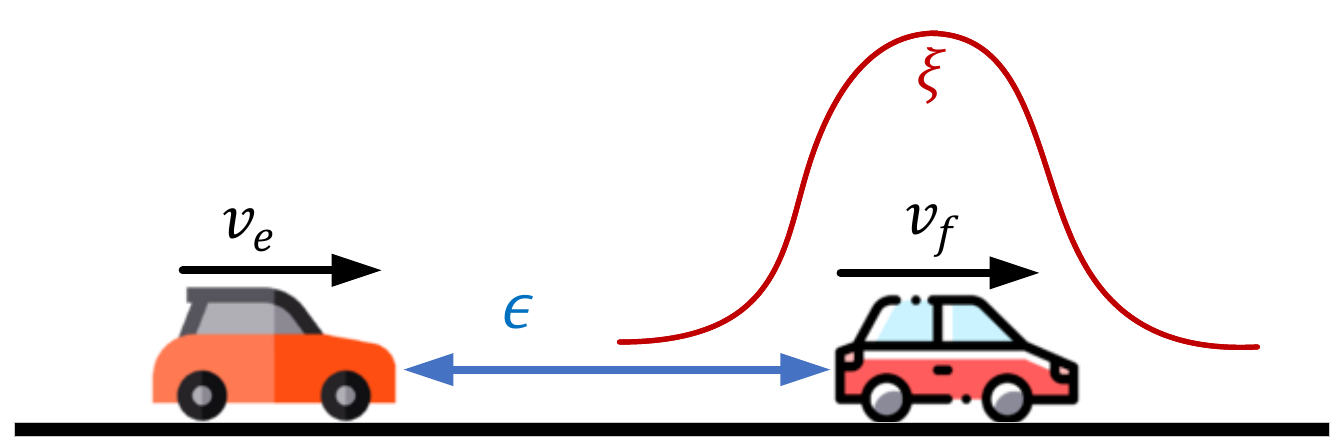}}
\caption{Car-following scenario.}
\label{fig:car-follwoing}
\end{figure}

In this section, the proposed SPIL is verified and compared in a car-following simulation in Fig. \ref{fig:car-follwoing}, where the ego car expects to drive fast and closely with the front car to reduce wind drag \cite{gao2016robust}, while keeping a minimum distance between the two cars with a high probability. Concretely,  we assume the ego car and front car follow a simple kinematics model, and the velocity of the front car is varying with uncertainty $\xi$.

\begin{figure*}[h]
    \centering
    \subfigure[Cumulative reward under $90.0\%$ threshold]{
        \includegraphics[width=0.3\textwidth]{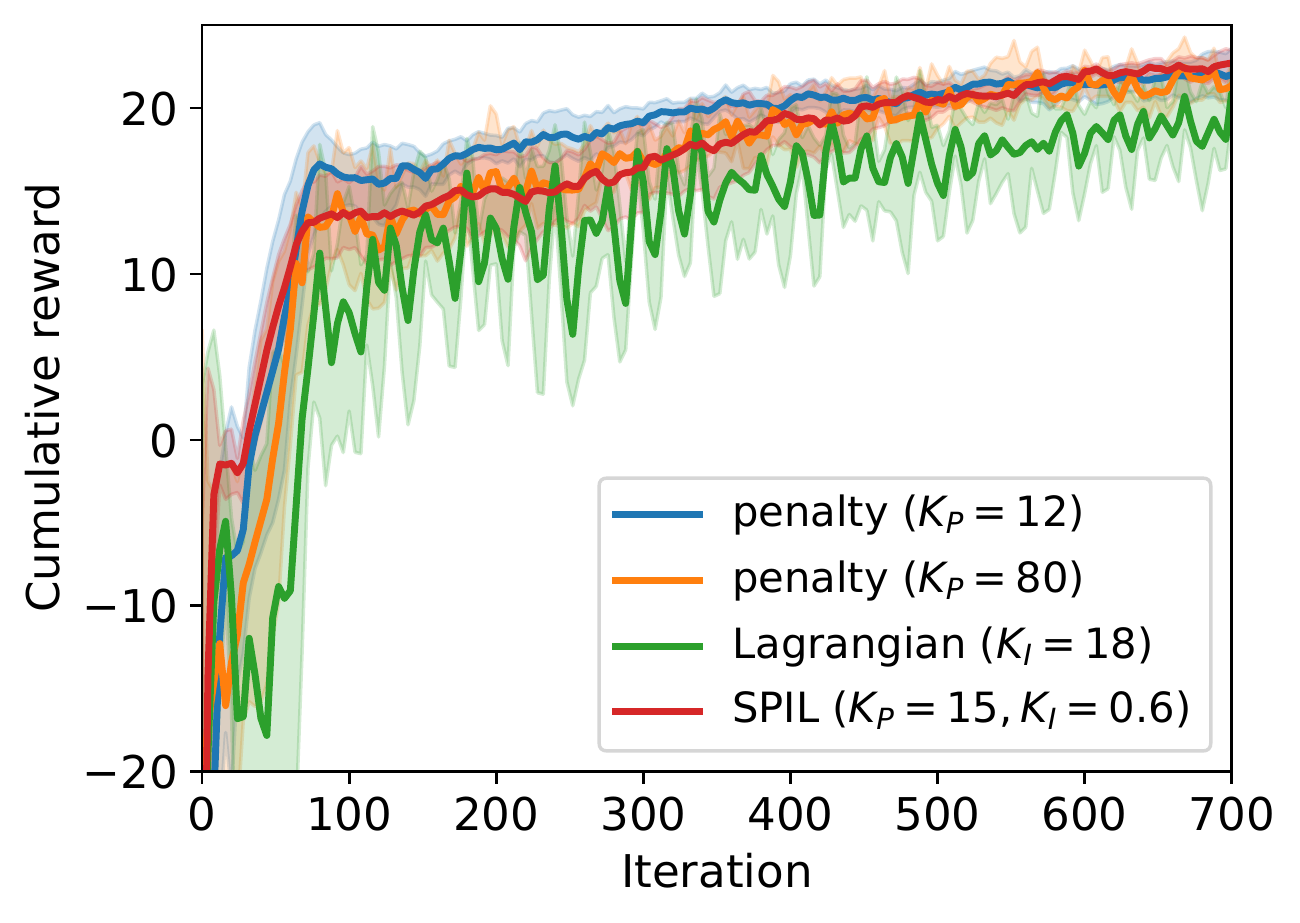}
        \label{return0.9}
    }
    \subfigure[Cumulative reward under $99.9\%$ threshold]{
	\includegraphics[width=0.3\textwidth]{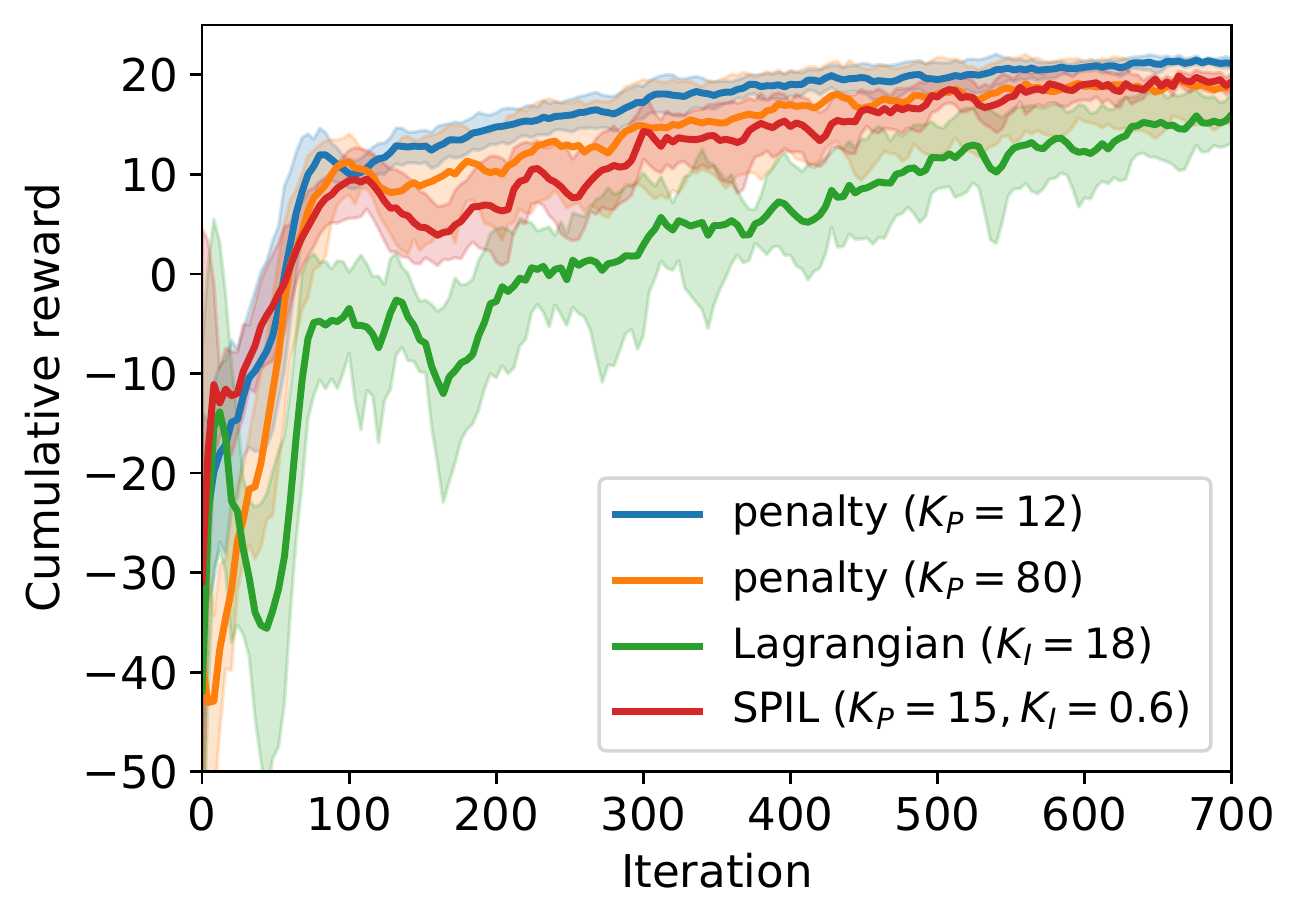}
        \label{return0.999}
    }
    \\   
    \subfigure[Safe probability under $90.0\%$ threshold]{
    	\includegraphics[width=0.3\textwidth]{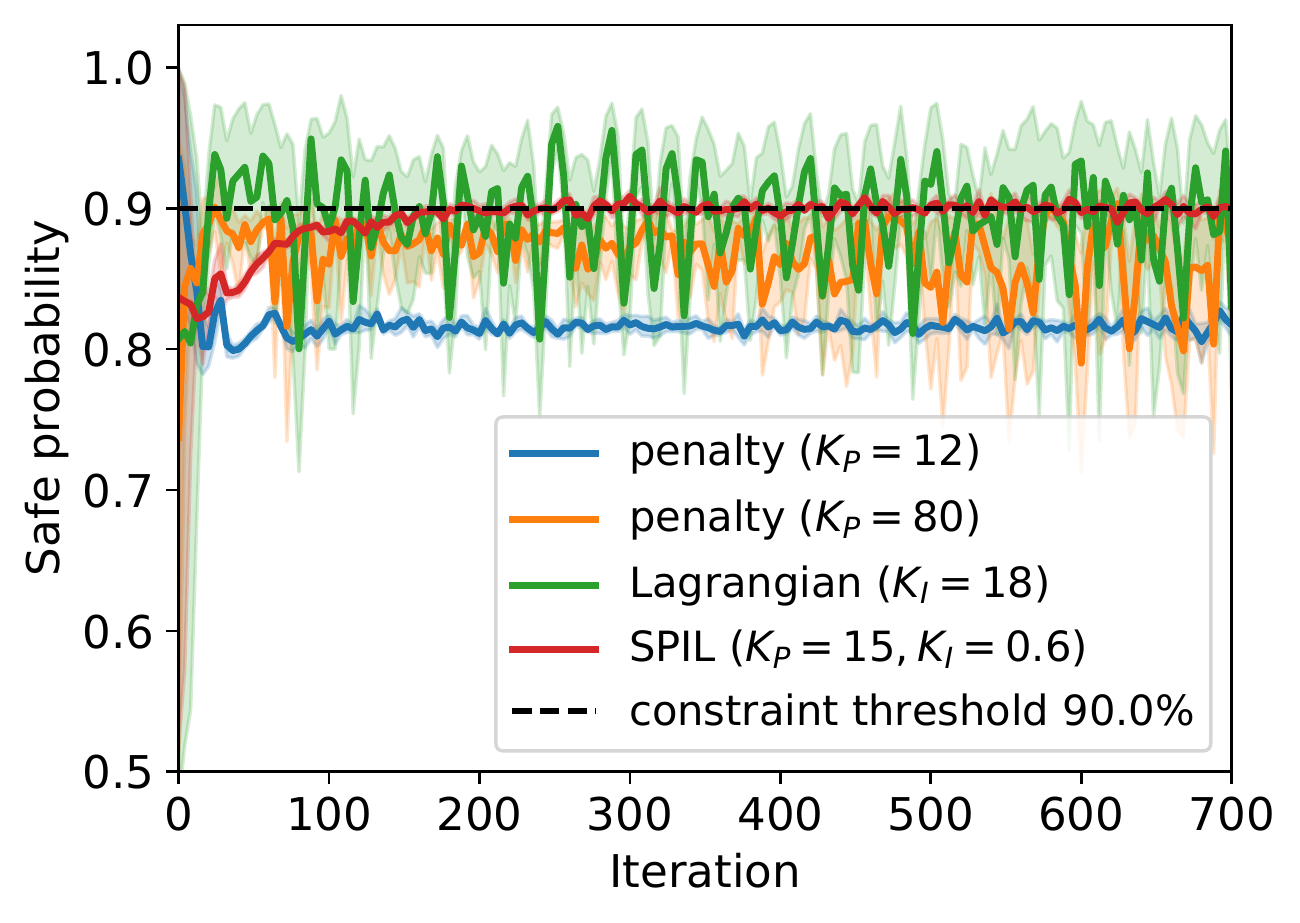}
        \label{safe0.9}
    }
    \subfigure[Safe probability under $99.9\%$ threshold]{
	\includegraphics[width=0.3\textwidth]{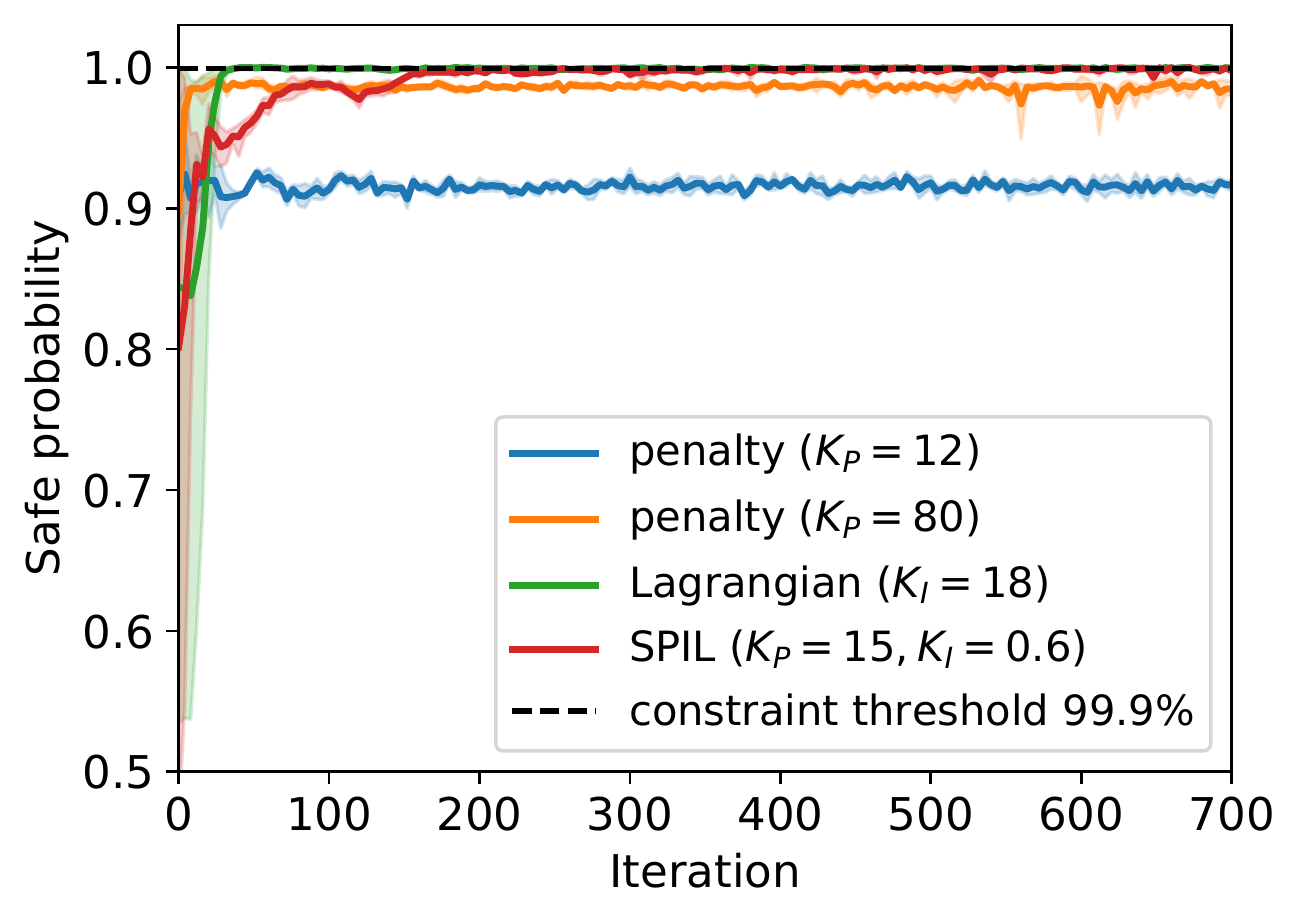}
        \label{safe0.999}
    }

    \caption{Comparison of performance among SPIL (separated Proportional-Integral Lagrangian), the penalty method and the Lagrangian method. The solid lines correspond to the mean and the shaded regions correspond to 95\% confidence interval over 5 runs.}
    \label{performance}
\end{figure*}

The dynamics of the car-following example is given as:
\begin{equation}
\begin{aligned}
&s_{t+1}=A s_t+B a_t+D \xi_t, \\
&s =[v_e,\quad v_f, \quad \epsilon]^{\top}, \\
&A=\left[{{\begin{array}{ccc}
1 & 0 & 0 \\
0 & 1 & 0 \\
-T & T & 1
\end{array}}}\right],\\
&B=[T,0,0]^{\top}, \quad D=[0,T,0]^{\top}
\end{aligned}
\end{equation}
where $v_e$ (m/s) is the velocity of ego car, $v_f$ (m/s) is the velocity of front car,
and $\epsilon$ (m) is the distance between the two cars. The action $a \in (-4,3) {\rm m/s^2}$ is the acceleration of ego car. The uncertainty $\xi_t \sim \mathcal{N}(0,0.7)$ is truncated in the interval $(-7,7)$. $T=0.1 s$ is the simulation time step. With a chance constraint on the minimum distance, the policy optimization problem is defined as:
\begin{equation}\begin{aligned}
\max _{\pi} &\sum_{t=0}^{\infty} \gamma^{t}(0.2v_{e,t}-0.1\epsilon_t-0.02a^2_t) \\
\text { s.t. } &{\rm Pr}\left\{ \bigcap_{t=1}^{N} \left(\epsilon_{t}>2\right)\right\}\ge1-\delta, \\
&s_{t+1}=A s_t+B a_t+D \xi_t
\end{aligned}\end{equation}
where $v_{e,t}$ denotes the velocity of the ego car at step $t$.

\subsection{Algorithm Details}
Three algorithms are employed to find the nearly optimal car-following policy, including SPIL (ours), the penalty method (amounts to proportional-only PIL), and the Lagrangian method (amounts to integral-only PIL). Note that all the algorithms are trained in the model-based manner. The coefficients of SPIL are selected as $K_P=15$, $K_I=0.6$ since it achieves the best results. The penalty method is sensitive to the penalty weight selection, so we adopt two weights $K_P=12$ and $80$. Actually, both of them cannot totally ensure the constraint satisfaction. For the Lagrangian method, we set $K_I=18$ because it achieved the best performance in the pre-test compared to other values. The cumulative reward and safe probability in horizon $N$ are compared under two chance constraint thresholds: 90.0\% and 99.9\%, i.e., $\delta=0.1$ and $\delta=0.001$. 

Both actor and critic are approximated by fully-connected neural networks. Each network has two hidden layers using rectified linear unit (ReLU) as activation functions, with 64 units per layer. The optimizer for the networks is Adam \cite{lecun2015deep}. The main hyper-parameters are listed in Table \ref{tab:hyper}.

\begin{table}[hbt!]
\caption{SPIL Hyper-parameters for Simulations}
\begin{center}
\label{tab:hyper}

\begin{tabular}{cc}
\hline
Parameters                       & Symbol                \& Value    \\ \hline
trajectories number             & $M=4096$       \\
constraint horizon              & $N=40$      \\
discount factor              & $\gamma=0.99$      \\
learning rate of policy network & $\alpha_{\theta}=3e-4$         \\
learning rate of value network  & $\alpha_{\omega}=2e-4$     \\ 
parameters of $K_S$             & $(\beta,\varepsilon_1,\varepsilon_2)=(0.3,0.2,0.05) $  \\
parameters of $\phi(\cdot)$     & $(\tau, b_1, b_2)=(1e-3, 1, 0.45)$                   \\
\hline
\end{tabular}
\end{center}
\end{table}

%Since $K_P=80$ already has serious oscillation, we do not choose a larger weight. 

% \begin{figure*}[htb!]
%     \centering
%     \subfigure[Safe probability]{
%         \includegraphics[width=0.3\textwidth]{unsafe_safety0999.pdf}
%         \label{unsafe_safety0.999}
%     }
%     \subfigure[Cumulative reward]{
% 	\includegraphics[width=0.3\textwidth]{unsafe_return0999.pdf}
%         \label{unsafe_return0.999}
%     }
%     \subfigure[Integral value]{
%     	\includegraphics[width=0.3\textwidth]{unsafe_integral0999.pdf}
%         \label{unsafe_integral0.999}
%     }
%     \caption{Comparison of performance of SPIL with and without integral separation under $99.9\%$ threshold. \textcolor{red}{The solid lines correspond to the mean and the shaded regions correspond to 95\% confidence interval over \textcolor{blue}{X} runs.}}
%     \label{unsafe_performance}
% \end{figure*}

\subsection{Results}
\subsubsection{Overall Performance}
The learning curves of all methods under two thresholds are illustrated in Fig. \ref{performance}. We emphasize that any comparison should consider both safety and reward-winning. Generally, the proposed SPIL algorithm not only succeeds to satisfy the chance constraint without periodic oscillations, but also achieves the best cumulative reward among methods that meet the safety threshold.

For safety, the proposed SPIL satisfies the chance constraint in both thresholds as shown in Fig. \ref{safe0.9} and Fig. \ref{safe0.999}. While the penalty methods with two weights both fail to meet the thresholds. Additionally, the large weight of $K_P=80$ also leads to oscillation. The Lagrangian method basically satisfies the constraint, but suffers from periodic oscillations under $90.0\%$ threshold. In a feedback control view, the SPIL combines the advantages of integral and proportion control, thus having a stable learning process with no steady-state errors. 

In terms of reward-wining plotted in Fig. \ref{return0.9} and Fig. \ref{return0.999}, SPIL achieves more reward than the Lagrangian method but less than the penalty method with $K_P=12$. This is because that the penalty method actually wins high performance at the cost of constraint violation. 

\subsubsection{Ablation Study}
\begin{figure}[hbt]
    \centering
    \subfigure[Safe probability]{
        \includegraphics[width=0.2057\textwidth]{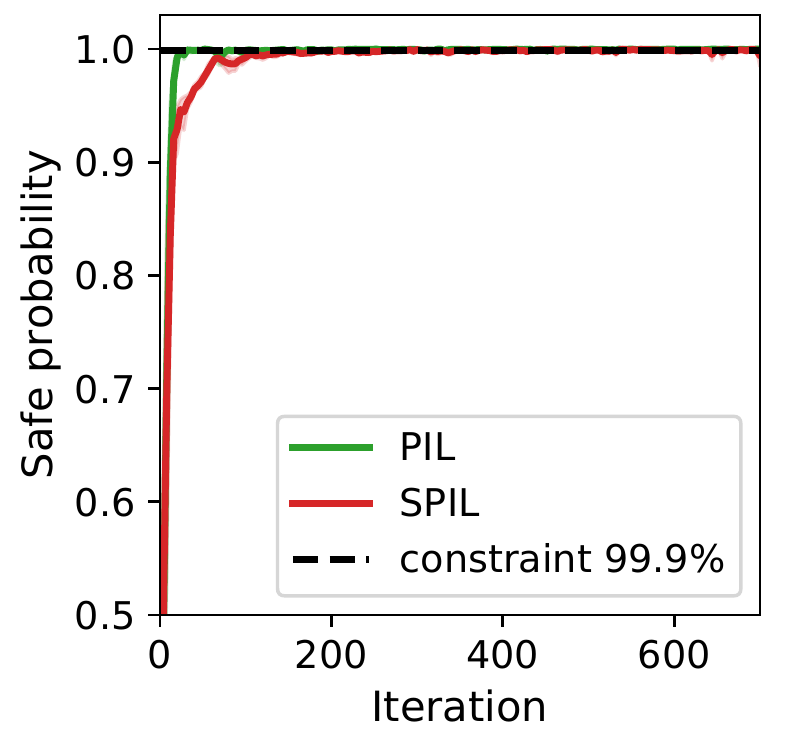}
        \label{unsafe_safety0.999}
    }
    \subfigure[Cumulative reward and integral value]{
	\includegraphics[width=0.2443\textwidth]{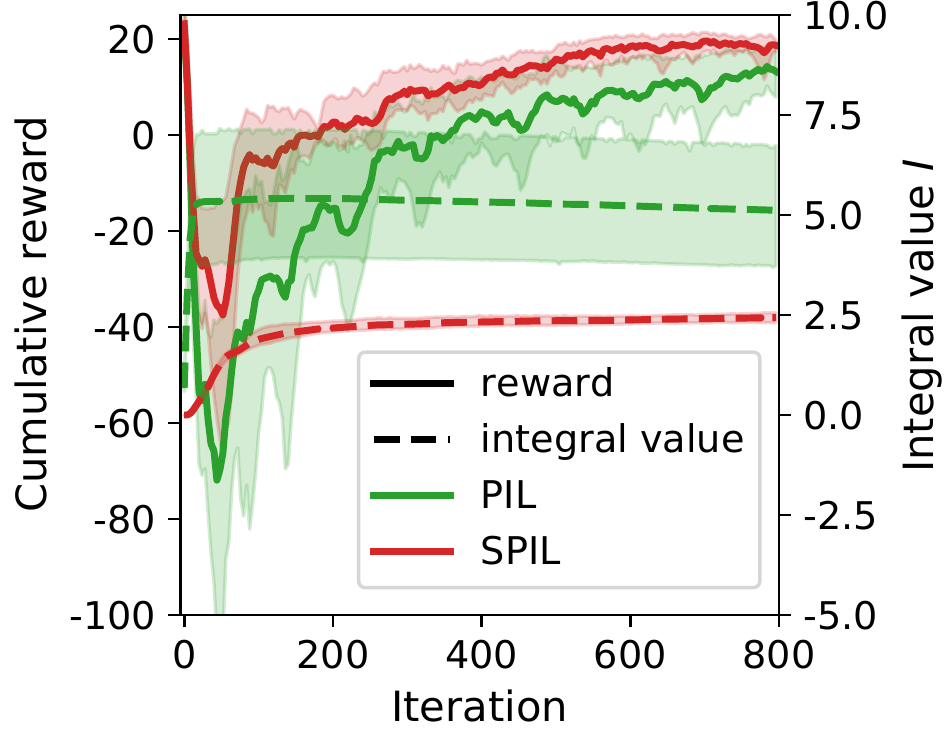}
        \label{unsafe_return0.999}
    }
    \caption{Comparison of performance of SPIL with and without integral separation technique (PIL) under $99.9\%$ threshold. The solid and dotted lines correspond to the mean and the shaded regions correspond to 95\% confidence interval over 5 runs. In (b), solid lines represent reward, dotted line represent integral value. The values for SPIL are in red and the values for PIL are in green. }
    \label{unsafe_performance}
\end{figure}

To demonstrate the necessity of integral separation proposed in Section \ref{separate}, we compare the results of our method with and without integral separation with five unsafe initial policies. As shown in Fig. \ref{unsafe_performance}, since the initial safe probability is low, the integral value $I$ increases rapidly at first. With a large $I$ and $\lambda$, the policy quickly becomes 100\% safe. Unfortunately, since $\Delta=0.999-1=-0.001$ is too small in \eqref{I_update}, the decline of $I$ and $\lambda$ is quite slow, leading to an overly conservative policy with poor reward. On the contrary, once the integral separation is equipped, the above problem is immediately solved. Note that the results in Fig. \ref{performance} and Fig. \ref{unsafe_performance} are not comparable since the latter are conducted under manually chosen unsafe initial policies.

\begin{table*}[hb!]
\caption{Algorithm Performance with different parameters}
\begin{center}

\begin{tabular}{cccccc}
\toprule
Value of $K_P$ ($K_I=0.6$, $\beta=0.3$)    & 3.75      & 7.5     & \textbf{15}     & 30   & 60                \\ \hline
Cumulative reward   & $22.27\pm0.56$     & $21.86\pm0.67$     & \bm{$23.01\pm0.96$}   & $21.91\pm1.60$  & $21.49\pm1.61$          \\ 
Safe probability   & $89.5\pm0.1\%$     & $90.2\pm0.3\%$     & \bm{$90.0\pm1.8\%$}   & $91.6\pm1.8\%$  & $88.2\pm6.6\%$           \\ 
\toprule
Value of $K_I$ ($K_P=15$, $\beta=0.3$)     & 0.15      & 0.3     & \textbf{0.6}     & 1.2   & 2.4       \\ \hline
Cumulative reward   & $21.60\pm0.66$     & $22.18\pm0.68$     & \bm{$23.01\pm0.96$}   & $21.77\pm0.43$  & $21.87\pm0.50$          \\ 
Safe probability   & $89.9\pm1.0\%$     & $89.0\pm0.4\%$     & \bm{$90.0\pm1.8\%$}   & $90.5\pm0.2\%$  & $89.5\pm0.3\%$           \\
\toprule
Value of $\beta$  ($K_P=15$, $K_I=0.6$)    & 0.1      & 0.2     & \textbf{0.3}     & 0.4   & 0.5             \\ \hline
Cumulative reward   & $21.88\pm0.90$     & $21.25\pm0.44$     & \bm{$23.01\pm0.96$}   & $21.86\pm1.44$  & $21.74\pm0.68$          \\ 
Safe probability   & $90.0\pm0.7\%$     & $90.8\pm1.0\%$     & \bm{$90.0\pm1.8\%$}   & $89.9\pm1.1\%$  & $90.1\pm0.4\%$           \\ \bottomrule
\end{tabular}
\end{center}
\label{tab:KP}
\end{table*}

\begin{figure}[hbt]
    \centering
    \subfigure[Safe probability]{
        \includegraphics[width=0.225\textwidth]{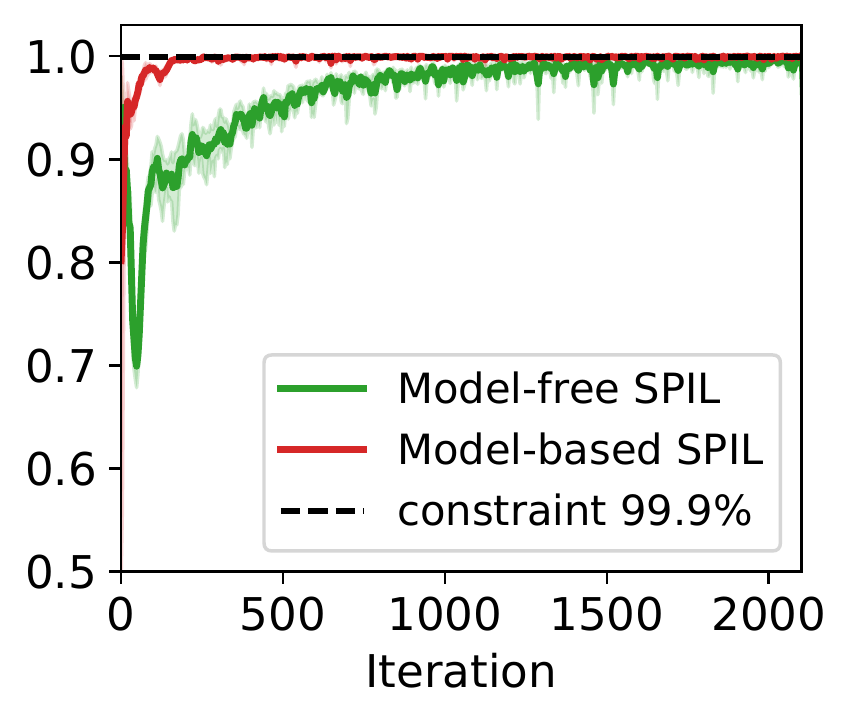}
        \label{pposafety0999}
    }
    \subfigure[Cumulative reward]{
	\includegraphics[width=0.225\textwidth]{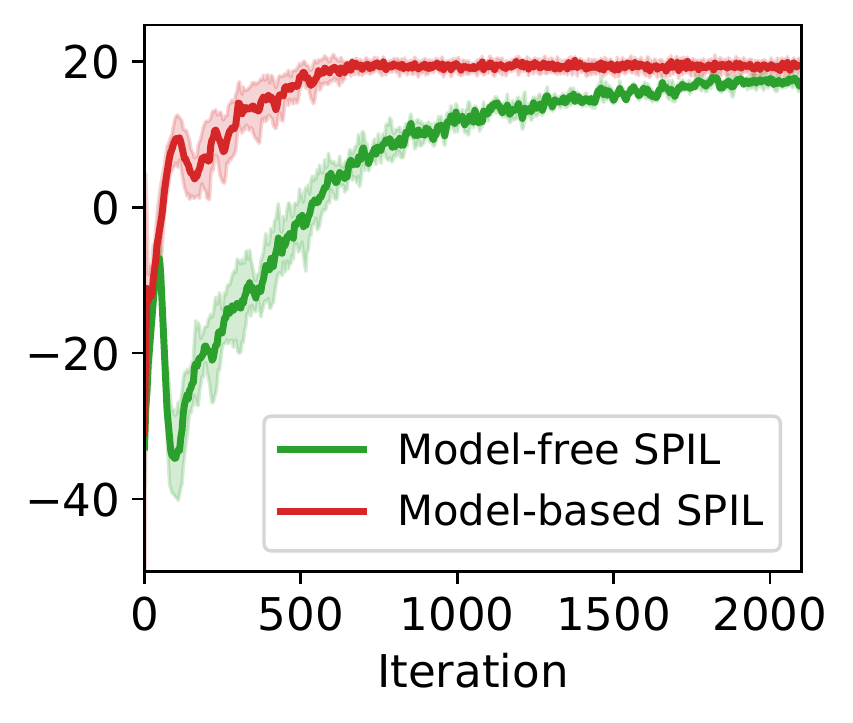}
        \label{pporeturn0999}
    }
    \caption{Comparison of performance of SPIL with model-based and model-free optimization  under $99.9\%$ constraint threshold. The solid lines correspond to the mean and the shaded regions correspond to 95\% confidence interval over 5 runs.}
    \label{ppo_performance}
\end{figure}

The previous works typically adopt model-free methods to optimize the policy due to the discontinuity of the indicator function. In Section \ref{model-based gradient}, we propose a model-based gradient of safe probability to enable model-based optimization, which is believed to accelerate the training process. To verify this, we use a popular model-free algorithm PPO \cite{Schulman2017ProximalPO} to compute the gradient of the safe probability, following \cite{Paternain2019LearningSP}. The learning curves of safe probability $p_s$ and cumulative reward $J$ are plotted in Fig. \ref{ppo_performance}, where an iteration amounts to a batch of 4096 state-action pairs for both model-based and model-free versions. Our model-based SPIL reaches the required safe probability within about 250 iterations, at least five times faster than its model-free counterpart. This improvement comes from the fact that the model-free method has first to learn an accurate cost value function before it optimizes the policy, while the model-based method makes use of the model to directly obtain a relatively precise gradient. It should be pointed out that this improvement is especially helpful for online training in the real world, where the policy should be safe as early as possible.

\subsubsection{Sensitivity Analysis}
\label{sec:sensitivity}
 To demonstrate the practicality of SPIL, we show that its high performance is relatively insensitive to hyper-parameter choice. We test the algorithm across different values of $K_P$ , $K_I$ and $\beta$, while keeping all other parameters fixed. The results over 5 runs under $90.0\%$ threshold are summarized in Table \ref{tab:KP}, with the best parameters shown in bold. Even the worst case only leads to 1.8\% degradation in safe probability and 6.5\% degradation in cumulative reward.

\section{Experimental Validation}

\label{sec:Experiment of AGV Obstacle Avoidance}
\subsection{Experiment Description}
\begin{figure}[htb!]
\centerline{\includegraphics[width=0.3\textwidth]{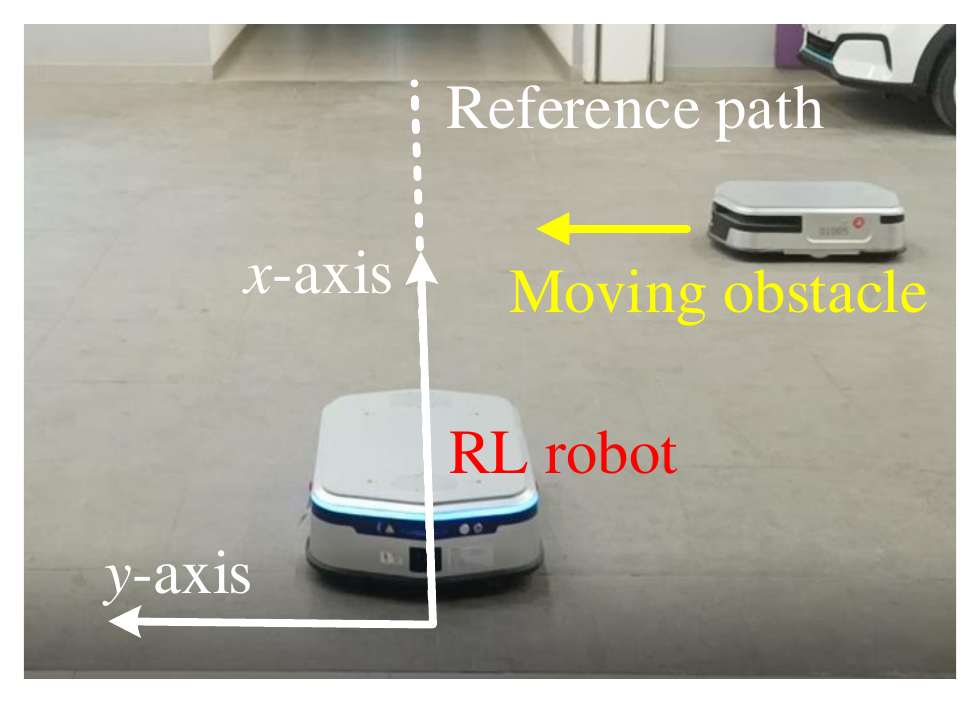}}
\caption{Mobile robot navigation task.}
\label{fig:AGV}
\end{figure}

To demonstrate the effectiveness of the proposed method for real-world safety-critical application, we apply it in a mobile robot navigation task. As illustrated in Fig. \ref{fig:AGV}, the robot aims to follow the reference path (exactly the positive $x$-axis) without colliding with a moving obstacle. However, it does not know the behaviour or trajectory of the moving obstacle, which may be highly stochastic. Besides, the moving obstacle will not actively avoid the robot. 

The robot locates its position and heading angle through a lidar, and it also has sensors to estimate its current velocity and angular velocity. In this experiment, we let the obstacle share its current motion information through socket communications. 
%We assume the AGV robot are also available to these motion information of the obstacle, which can be obtained through a perception module or communication module. 

\subsection{Experiment Details}

\subsubsection{Dynamic Model}
\label{sec:experiment model}
We first present the model used for training. The wheeled robot adopts a two-wheel differential drive architecture, and takes the desired velocity $v^d$ and desired angular velocity $w^d$ as control commands. Nonetheless, its bottom-level control mechanism and response are unknown and varied in different conditions. %Therefore, we assume the velocity $v$ and angular velocity $\omega$ are adjusted to their desired values in one step if their variations are small, but with 
Therefore, additional random terms $\xi^v$ and $\xi^\omega$ are introduced to describe this uncertainty. To cover the stochastic behaviour of obstacle, the uncertainty of the obstacle is also considered. The motions of both robot and moving obstacle are predicted through a simple kinematic model:
\begin{align}
 \begin{aligned}
    s&= 
\left[{{\begin{array}{ccccc}
P^x \\
P^y \\
\alpha \\
v \\
\omega
\end{array}}}\right],
a=\left[{{\begin{array}{cc}
v^d \\
w^d 
\end{array}}}\right], \\
s_{t+1}&=
\left[{{\begin{array}{ccccc}
P^x_{t} \\
P^y_{t} \\
\alpha_{t} \\
0 \\
0
\end{array}}}\right] +
\left[{{\begin{array}{ccccc}
0 \\
0\\
0 \\
v^d_{t} \\
\omega^d_{t} 
\end{array}}}\right]+ T
\left[{{\begin{array}{ccccc}
v_t\cos{\alpha} \\
v_t\sin{\alpha} \\
\omega_t\ \\
\xi^v_t \\
\xi^{\omega}_t 
\end{array}}}\right]
 \end{aligned}
 \end{align}
where $T=0.4s$ is the time step, $P^x$ and $P^y$ denote the position coordinates, $\alpha$ is the heading angle, $v$ is the velocity, $\omega$ is the angular velocity, $v^d$ is desired velocity, $w^d$ is desired angular velocity, $\xi^v$ is the uncertainty on the velocity, $\xi^\omega$ is the uncertainty on the angular velocity. For the robot, the control inputs $a$ are output by the policy network, which are also bounded by the following input constraints $|v-v^d|\le1.8T$, $ |\omega-\omega^d|\le0.8T$. The velocity and angular velocity uncertainty of the robot is set to $\xi^v \sim \mathcal{N}(0,0.08)$ and $\xi^{\omega} \sim \mathcal{N}(0,0.05)$. For the model of the obstacle, the desired velocity and angular velocity are always set as the same as its current velocity and angular velocity, but with uncertainty $\xi^v \sim \mathcal{N}(0,0.1)$ and $\xi^{\omega} \sim \mathcal{N}(0,0.06)$ to indicate its stochastic behaviors. Although the true future behavior of obstacles is unknown to the ego robot, these uncertainty terms help to improve the robustness of the learned policy in real environments.

% The uncertainty is chosen as $\xi^v \sim \mathcal{N}(0,0.08)$ and $\xi^{\omega} \sim \mathcal{N}(0,0.05)$ to indicate its unknown response. For the moving obstacle, the control signal $u$ is regarded the same as the velocity and angular velocity at last step, and the uncertainty is set as $\xi^v \sim \mathcal{N}(0,0.1)$ and $\xi^{\omega} \sim \mathcal{N}(0,0.06)$ to indicate its stochastic behaviours. This means that our model predicts obstacles will randomly and gradually change its speed and direction. This means that our model predicts obstacles will randomly and gradually change its speed and direction. 

We admit the whole model is relatively naive and inaccurate, and the uncertainty term is given by experience instead of estimation. However, we find it is enough to accomplish this task. A better choice is to update the model and uncertainty online through real-world experimental data. We leave it in our further work.

\subsubsection{Reward and Chance Constraints}
The robot aims to follow a reference path while avoiding collisions with a moving obstacle. The start point for the robot is near (1, 0). The reference path is the positive $x$-axis. Therefore, the reference $y$-position and heading angle are both 0. The reference velocity is set to 0.3 m/s. With additional regularization on the control command, the reward of the task is defined as:   
\begin{equation}
r=-1.4(P^y)^2-\alpha^2-16(v-0.3)^2-0.2(v^d)^2-0.5(\omega^d)^2
\end{equation}

Then we impose the obstacle avoidance constraint. For simplicity, the robot and obstacle are both regarded as circles with a radius of 0.4m, and the distance between their centers is denoted as $\epsilon$. The chance constraint on the minimum distance is:  
\begin{equation}
{\rm Pr}\left\{ \bigcap_{t=1}^{25} \left(\epsilon_{t}>0.9\right)\right\}\ge0.99
\end{equation}

\subsubsection{Algorithmic Parameters}
\label{sec:experiment parameters}
The network structure is exactly the same as that of the simulation. The main hyper-parameters are listed in Table \ref{tab:exp_hyper}.

\begin{table}[hbt!]
\caption{SPIL Hyper-parameters for Experiment}
\begin{center}
\label{tab:exp_hyper}

\begin{tabular}{cc}
\hline
Parameters                       & Symbol                \& Value    \\ \hline
trajectories number             & $M=4096$         \\
constraint horizon              & $N=25$              \\
learning rate of policy network & $\alpha_{\theta}=3e-2$       \\
proportional coefficient        & $K_P=60$            \\
integral coefficient            & $K_I=0.02$        \\
parameters of $K_S$             & $(\beta,\varepsilon_1,\varepsilon_2)=(0.7,0.2,0.1)$       \\
parameters of $\phi(\cdot)$     & $(\tau,b_1,b_2)=(7e-2,1,0.45 )$        \\
\hline
\end{tabular}
\end{center}
\end{table}

\subsubsection{Test Scenarios}
\label{sec:scenarios}
The learned policy will be tested in five scenarios with different obstacle behaviours. In the former three scenarios, the obstacle behaves normally, without a sudden stop or turn. To demonstrate the robustness and high intelligence of the trained robot, the latter two scenarios are under high randomness, where the obstacle drives in a complex trajectory or even deliberately blocks the robot. We stress that, in all experiments, the robot does not know the behavior or trajectory of the obstacle in advance, and all the experiments are conducted with the same network. This means that the method should have high generalization ability and intelligence to pass the five test scenarios. 

% \begin{table}[htb!]
% \caption{Testing scenarios}
% \begin{center}
% \begin{tabular}{ccc}
% \hline
% Scenario&Obstacle randomness    &Obstacle behavior                                              \\ \hline
% 1 &low           &Pass in a constant low speed                                           \\ 
% 2  &         &Pass in a constant high speed                                          \\ 
% 3   &        &Pass in a oblique line                                                \\ \hline
% 4 &high           &Pass in a sine curve                                             \\ 
% 5&                  & Block aggressively                                               \\ \hline
% \end{tabular}
% \end{center}
% \label{tab:scenario}
% \end{table}

\subsection{Results}
% \begin{figure*}[hb]
% \centerline{\includegraphics[width=0.9\textwidth]{sine_snap.pdf}}
% \caption{Snapshot for the scenario of obstacle in a sine curve.}
% \label{fig:sine_snap}
% \end{figure*}

% \begin{figure*}[hb]
% \centerline{\includegraphics[width=0.9\textwidth]{block_snap.pdf}}
% \caption{Snapshot for the scenario of obstacle blocking aggressively.}
% \label{fig:block_snap}
% \end{figure*}

\begin{figure*}[hbt]
    \centering
    \subfigure[Scenario 1]{
        \includegraphics[width=0.3\textwidth]{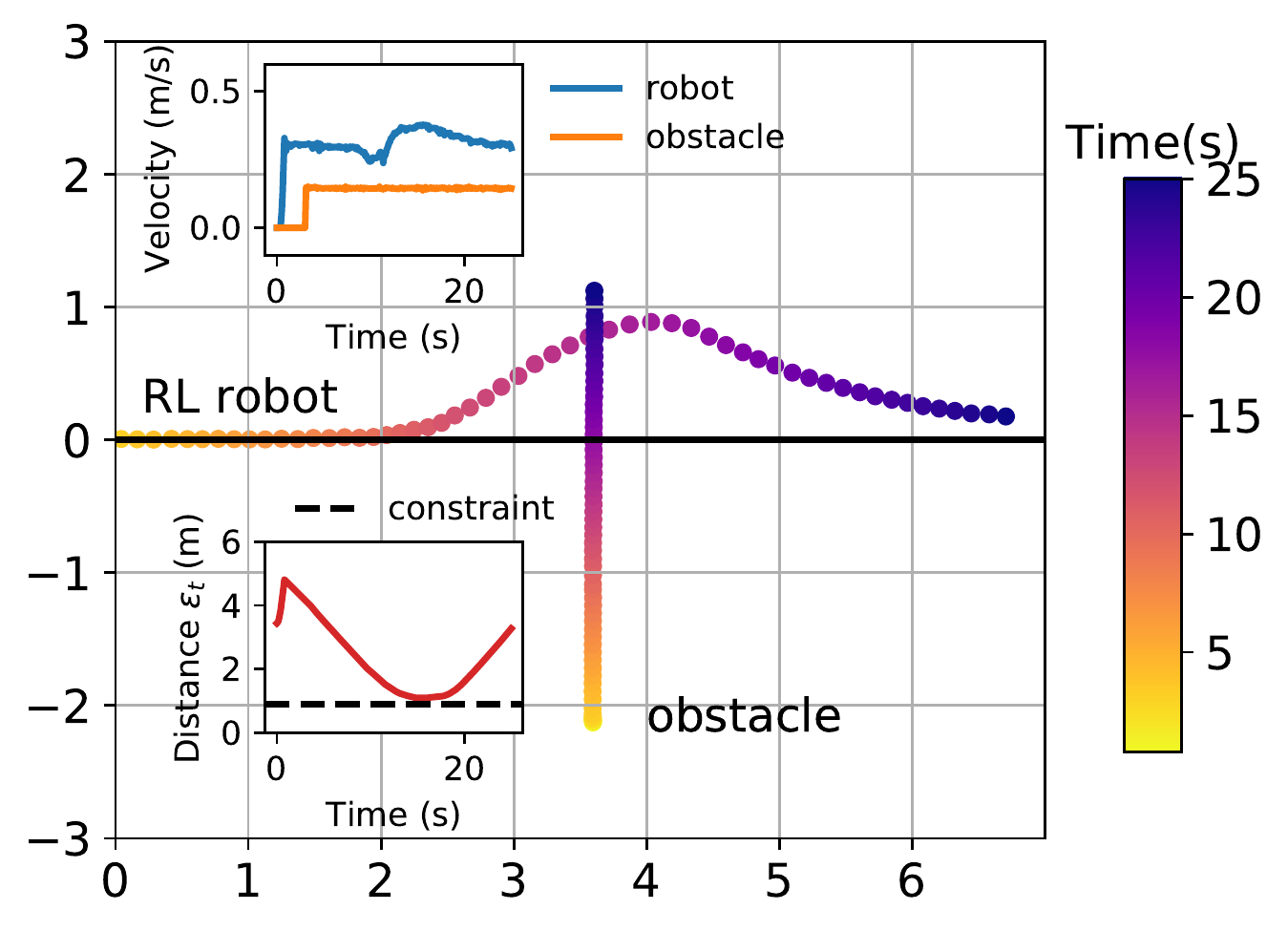}
        \label{fig:low speed}
    }
    \subfigure[Scenario 2]{
	\includegraphics[width=0.3\textwidth]{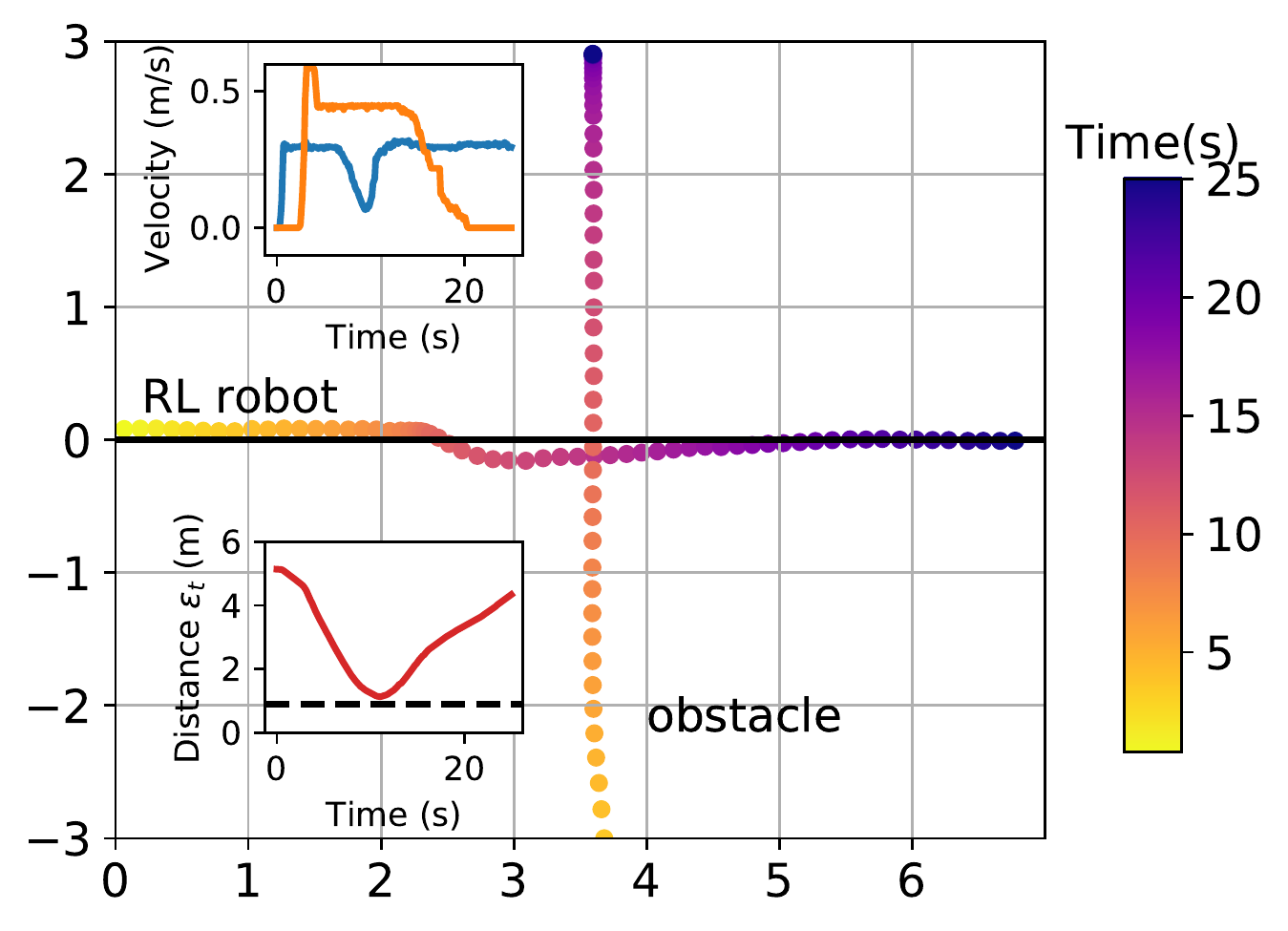}
        \label{fig:high speed}
    }
    \subfigure[Scenario 3]{
	\includegraphics[width=0.3\textwidth]{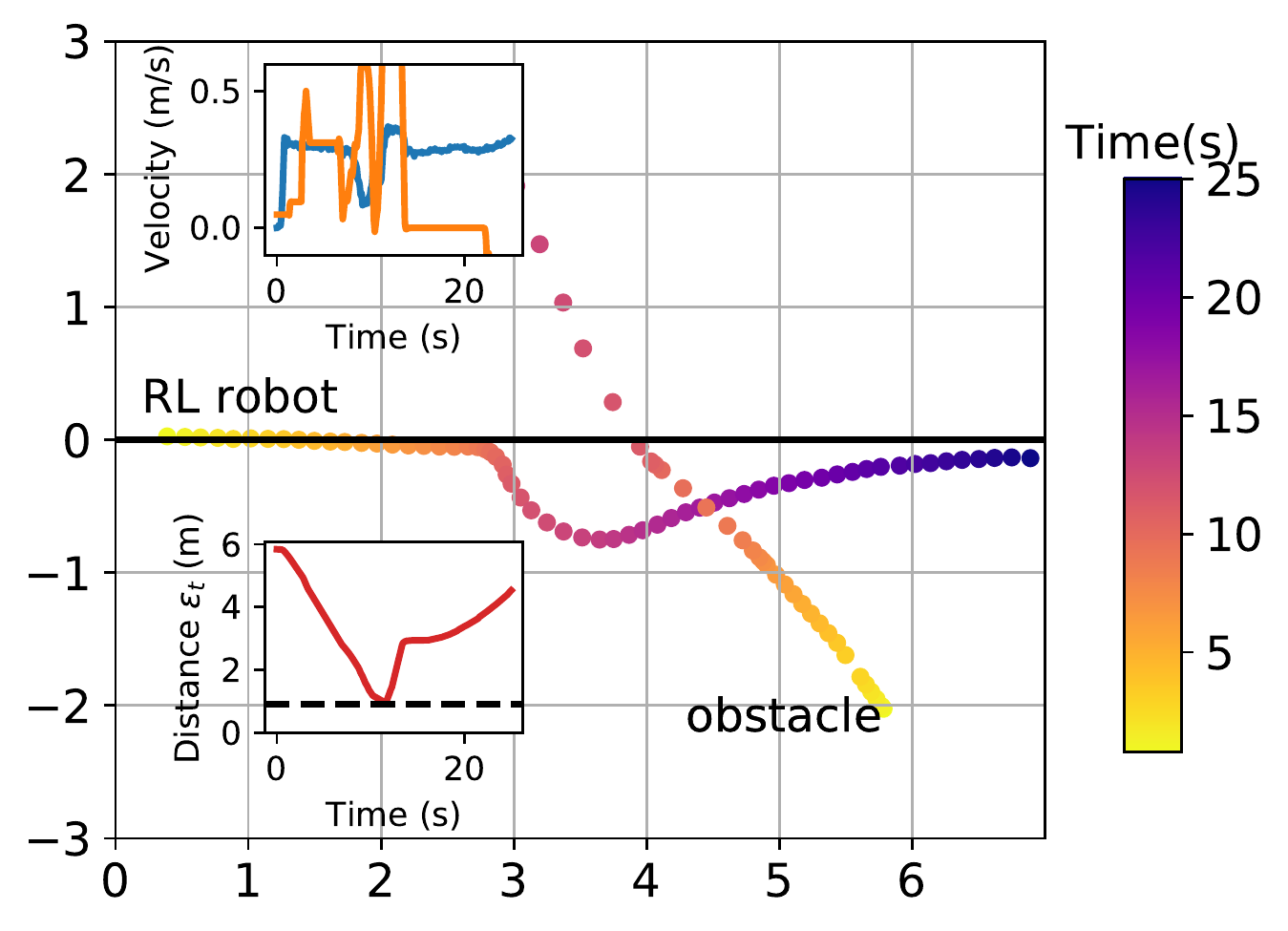}
    \label{fig:oblique line}
    }
    \subfigure[Scenario 4]{
	\includegraphics[width=0.3\textwidth]{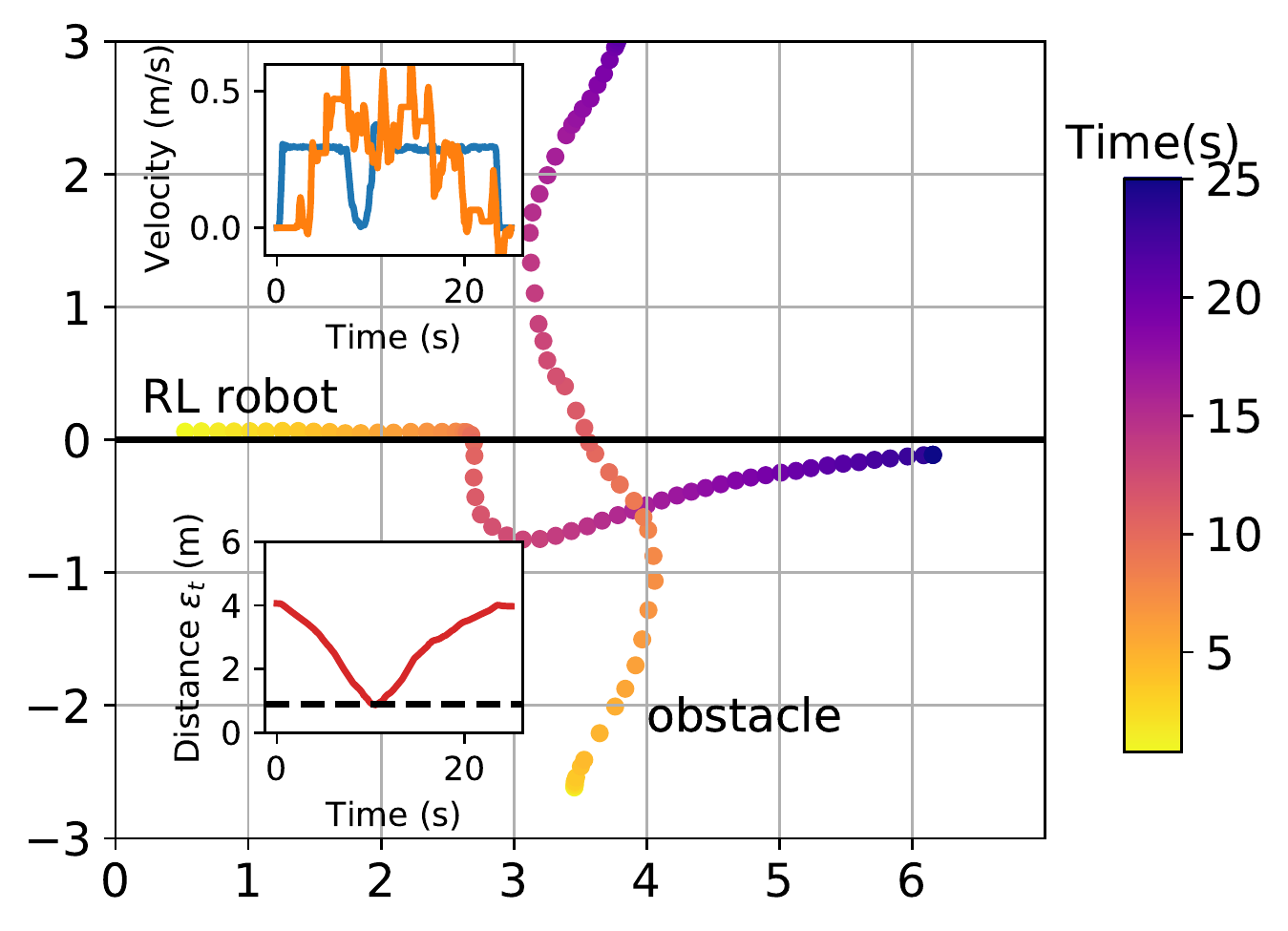}
    \label{fig:sine curve}
    }
     \subfigure[Scenario 5]{
	\includegraphics[width=0.3\textwidth]{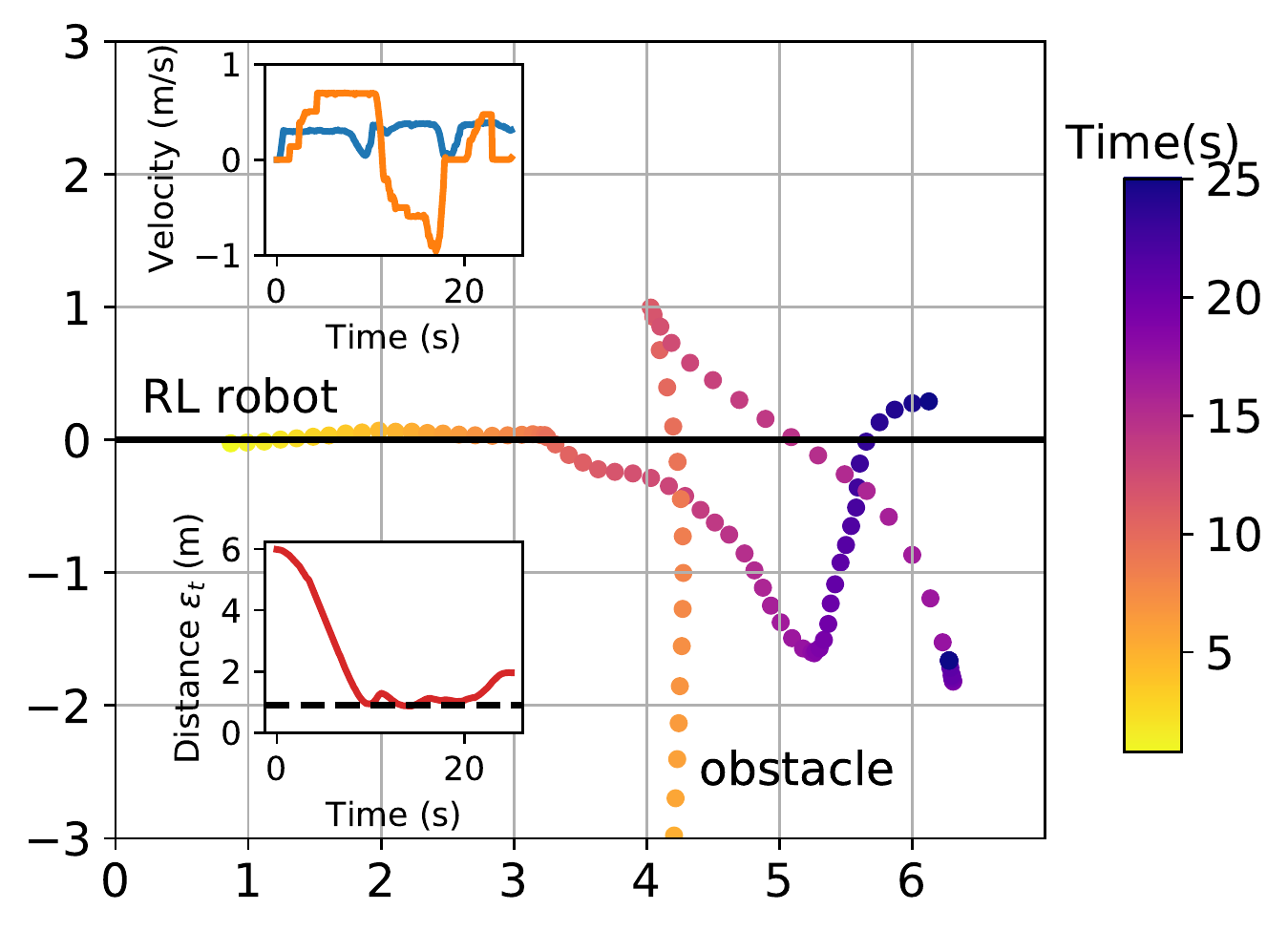}
    \label{fig:obstruct}
    }
    \caption{Trajectories of robot and obstacle in five scenarios. All points are plotted with the same time interval, so the density of points also represents the velocity of the object. The speed of two objects and distance between two objects are also plotted. In most time, our robot keeps a safe distance from the obstacle as specified in the chance constraint.}

\end{figure*}

We trained the policy network using the proposed  SPIL algorithm with the dynamic model, and then tested the learned network on the real robot. A video for the results is available at \href{https://youtu.be/oVDB2XqNoCU}{https://youtu.be/oVDB2XqNoCU}. We highly recommend readers to watch the video for an intuitive impression. Fig. \ref{unsafe_performance} shows the control results of the learned policy in five scenarios. In Fig. \ref{fig:low speed}, when the obstacle was moving at a low speed, the robot actively bypassed it from the left. If we increased the speed of the obstacle, the robot instead chose to stop and wait until the obstacle moved away, thereby avoiding a collision (See Fig. \ref{fig:high speed}). In Fig. \ref{fig:oblique line}, the ego robot avoided the moving obstacle from the right side while tracking the reference path. These results indicate that the learned policy can adopt different strategies according to the position and speed of the obstacle to achieve collision avoidance while tracking. 

% \begin{figure}[H]
% \centerline{\includegraphics[width=0.3\textwidth]{Obstacle passes with constant low speed.pdf}}
% \caption{Obstacle passes with a constant low speed.}
% \label{fig:low speed}
% \end{figure}

% \begin{figure}[H]
% \centerline{\includegraphics[width=0.3\textwidth]{Obstacle passes with constant high speed.pdf}}
% \caption{Obstacle passes with a constant high speed.}
% \label{fig:high speed}
% \end{figure}

% \begin{figure}[H]
% \centerline{\includegraphics[width=0.3\textwidth]{Obstacle passes in oblique line.pdf}}
% \caption{Obstacle passes in a oblique line.}
% \label{fig:oblique line}
% \end{figure}

\begin{figure*}[htb!]
    \centering
    \subfigure[Time:10s]{
        \includegraphics[width=0.22\textwidth]{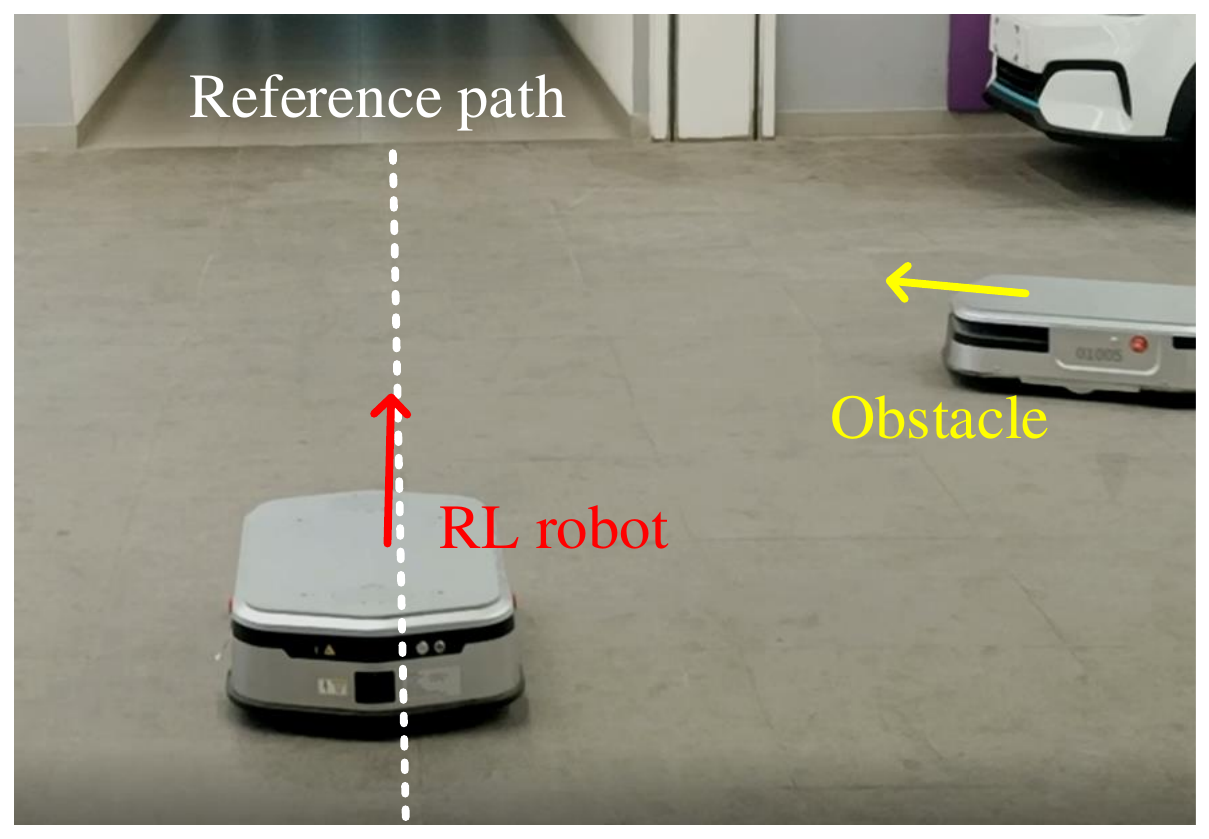}
    }
    \subfigure[Time:13s]{
        \includegraphics[width=0.22\textwidth]{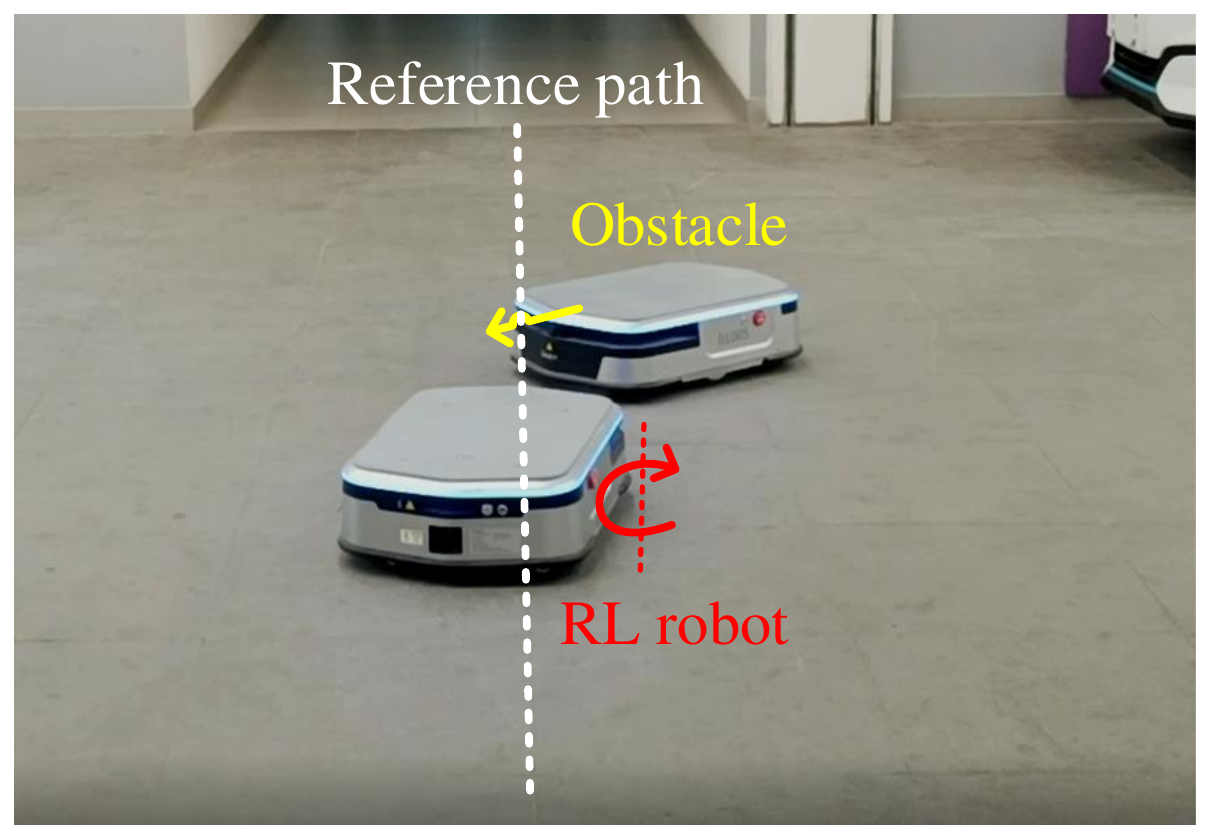}
    }
    \subfigure[Time:15s]{
        \includegraphics[width=0.22\textwidth]{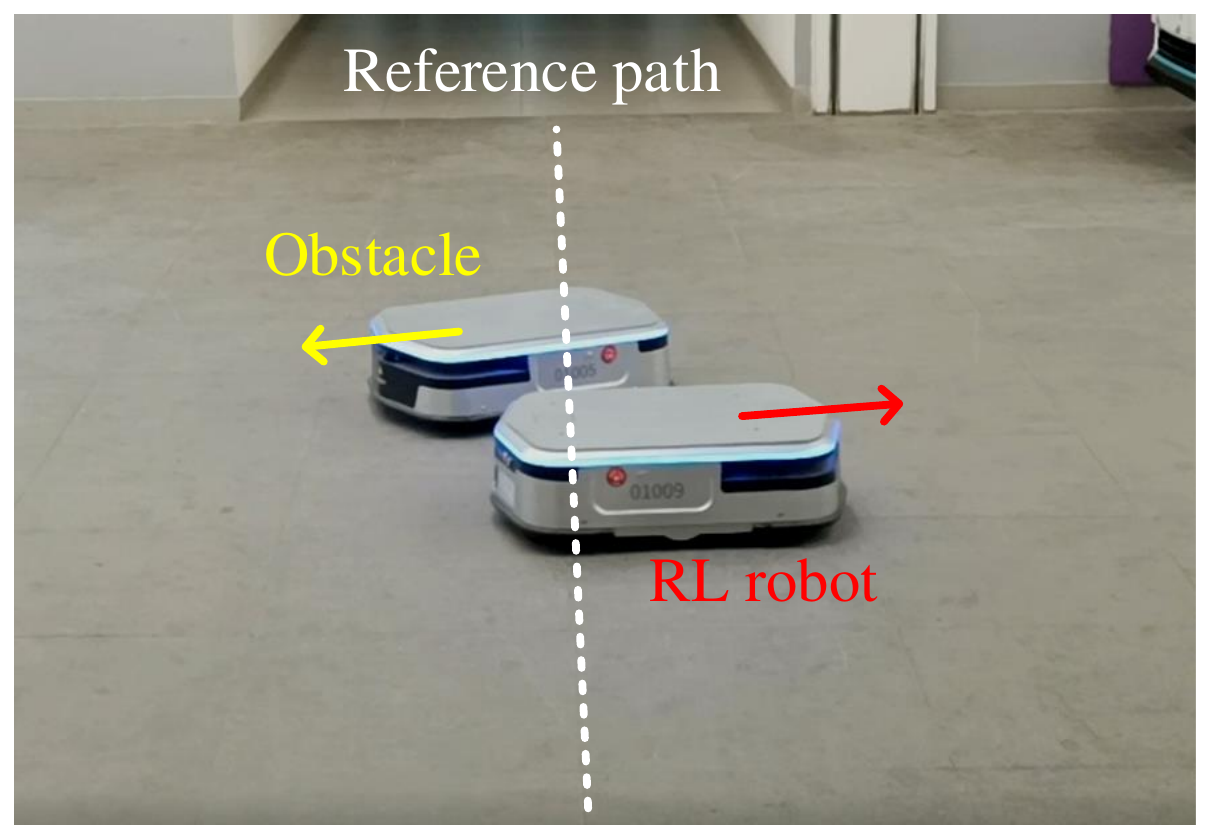}
    }
    \subfigure[Time:17s]{
        \includegraphics[width=0.22\textwidth]{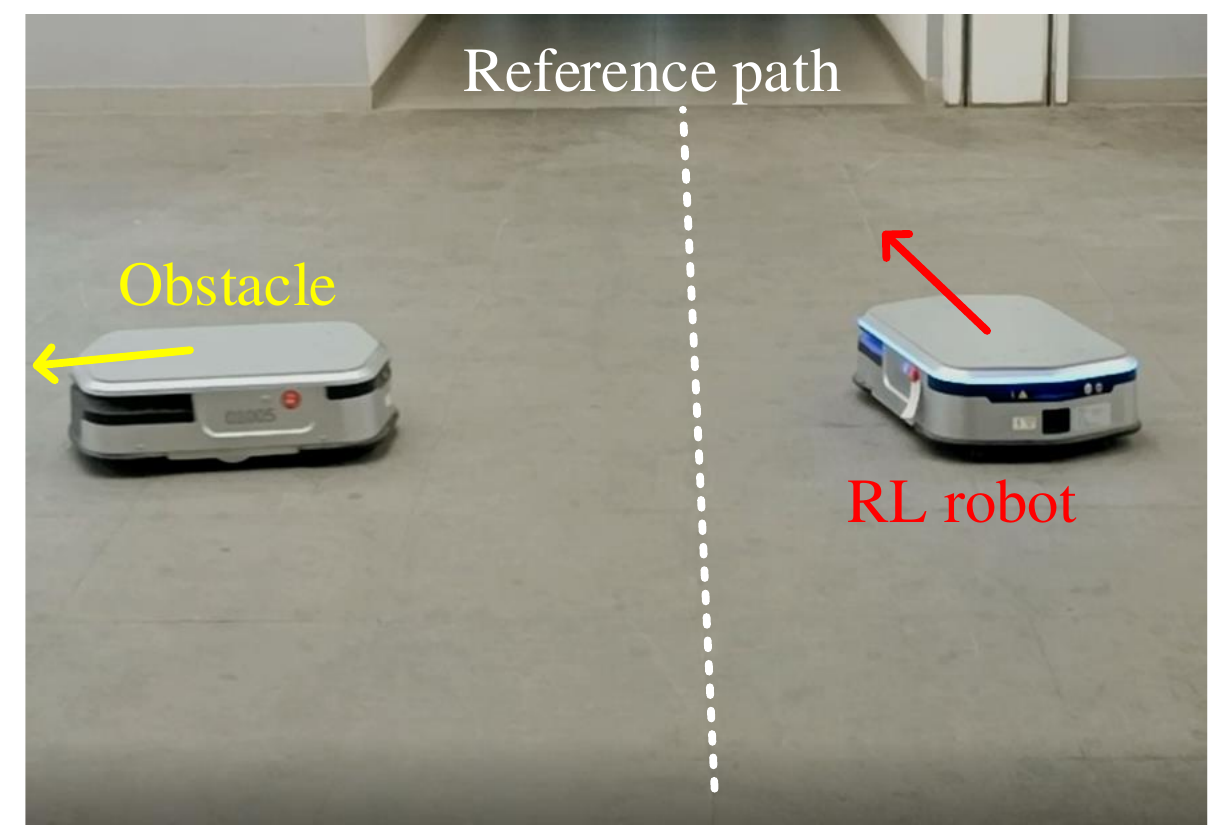}
    }
    \caption{Snapshots for the scenario 4. (a) The obstacle initially moved towards the positive direction of the reference path. (b) The obstacle suddenly changed its direction and moved towards the robot. Thus the robot urgently turned right to avoid it. (c) The robot passed around the obstacle from the right. (d) The robot returned to the reference path after the obstacle left.}
    \label{fig:sine_snap}
\end{figure*}

\begin{figure*}[htb!]
    \centering
    \subfigure[Time:11s]{
        \includegraphics[width=0.22\textwidth]{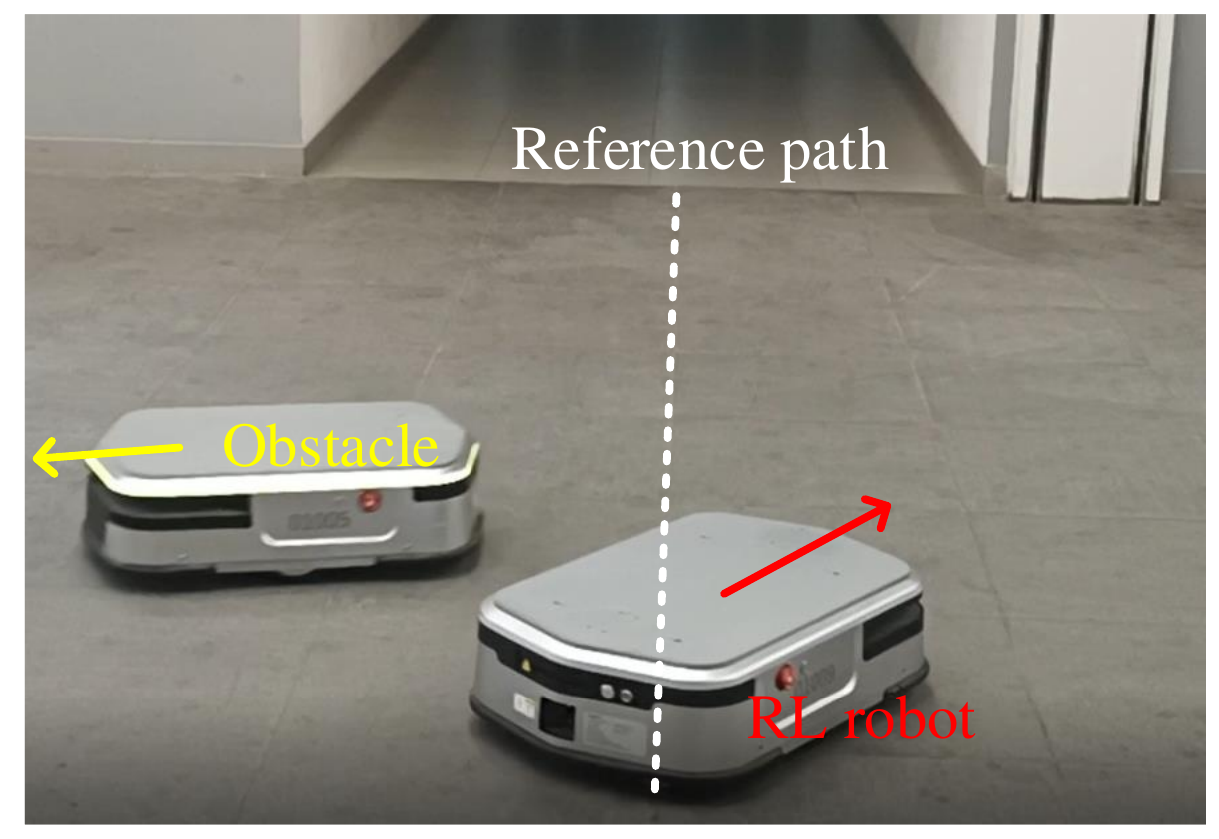}
    }
    \subfigure[Time:15s]{
        \includegraphics[width=0.22\textwidth]{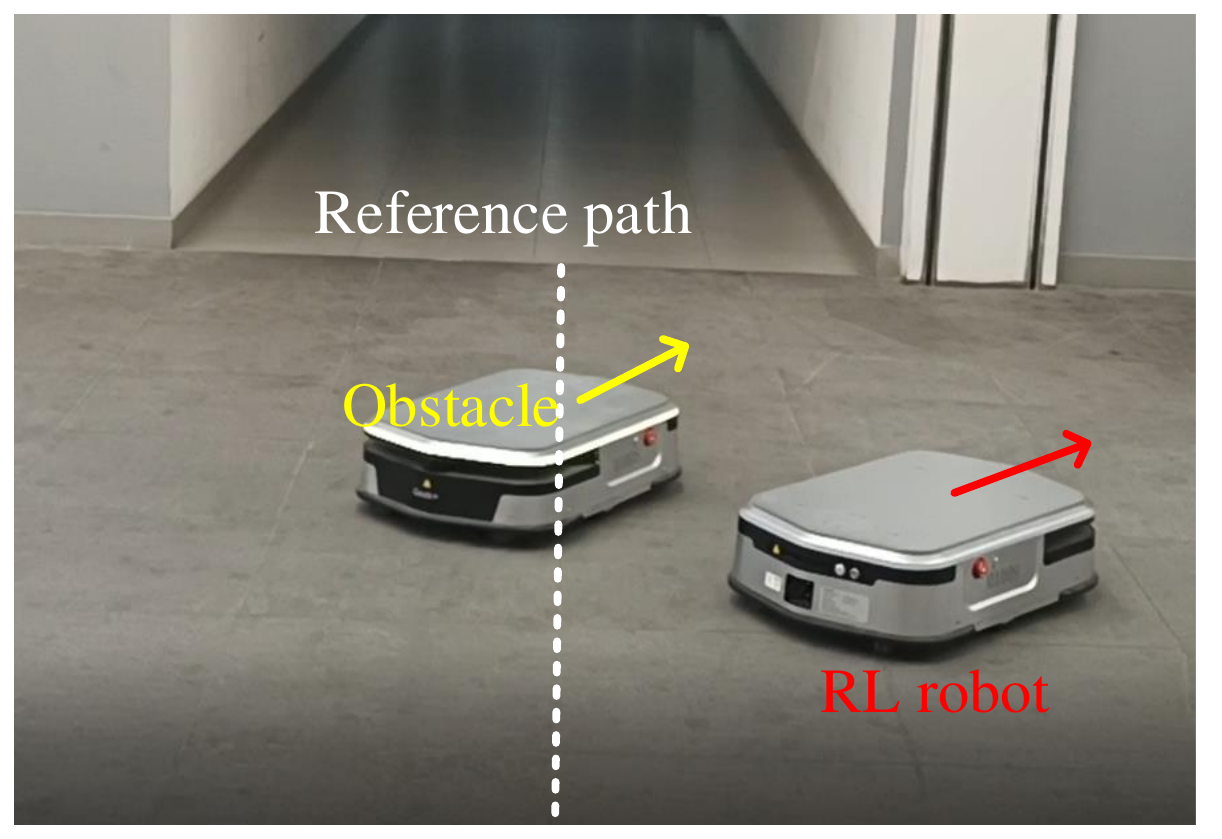}
    }
    \subfigure[Time:18s]{
        \includegraphics[width=0.22\textwidth]{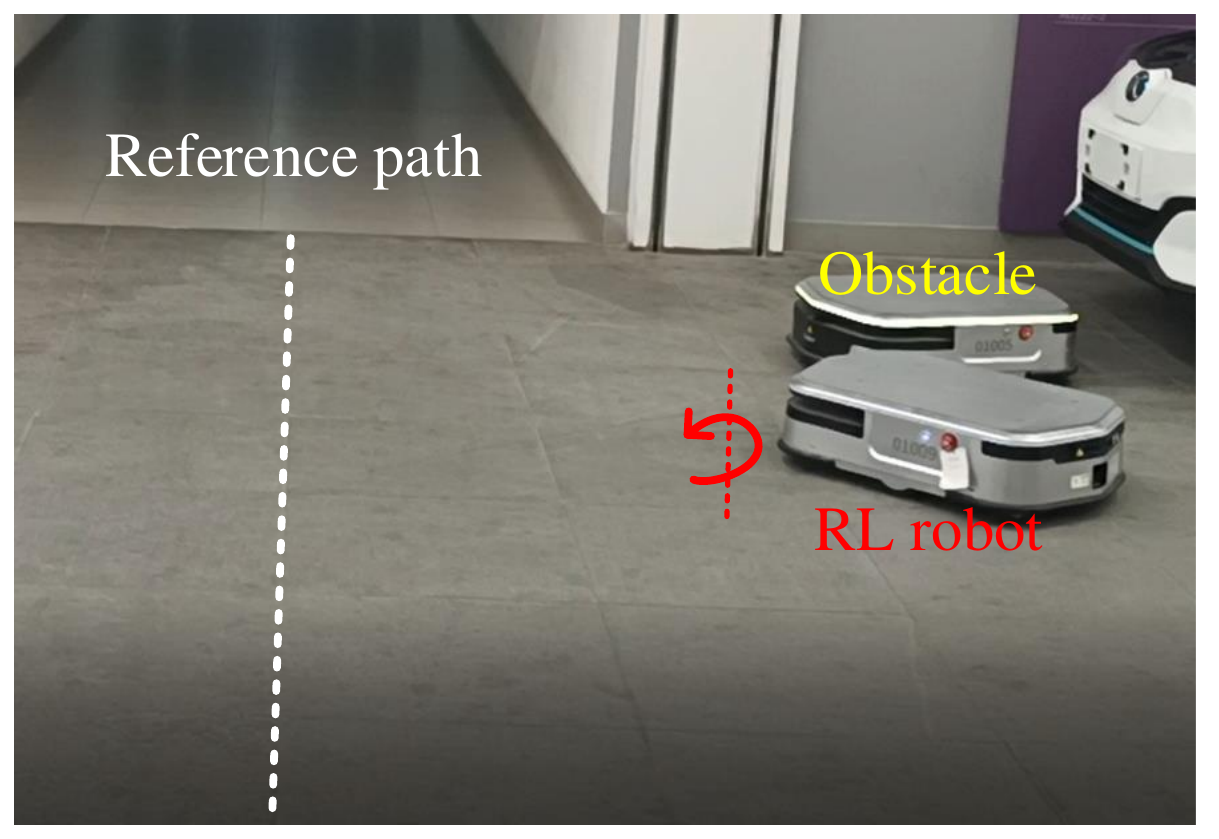}
    }
    \subfigure[Time:24s]{
        \includegraphics[width=0.22\textwidth]{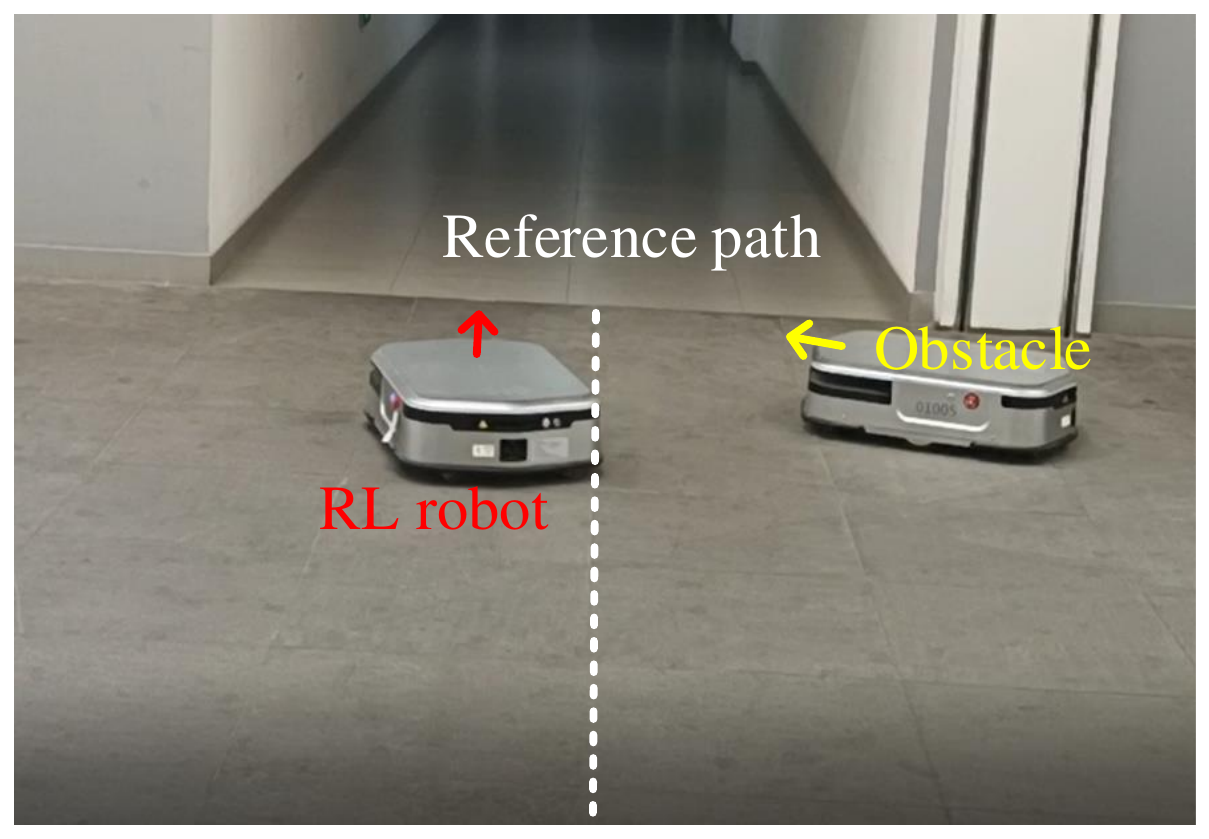}
    }
    \caption{Snapshots for the scenario 5. (a) The robot turned right to avoid the coming obstacle. (b) The obstacle deliberately moved back and blocked the robot from returning to the reference path. Thus the robot had to keep moving in the wrong direction. (c) The robot found an opportunity to turn left when the obstacle was not blocking. (d) The robot successfully returned to the reference path.}
    \label{fig:block_snap}
\end{figure*}

In scenarios 4 (Fig. \ref{fig:sine curve}) and 5 (Fig. \ref{fig:obstruct}), the behaviour of the obstacle robot was more aggressive. We controlled the obstacle to deliberately collide or block the  robot to increase the difficulty of collision avoidance. See Fig. \ref{fig:sine curve} and \ref{fig:sine_snap} for behavior details of the robot in scenario 4, and Fig. \ref{fig:obstruct} and \ref{fig:block_snap} for that in scenario 5. Results show that the learned policy can achieve safe movement even when the obstacle behaves aggressively, which demonstrates the amazing performance of our method.

% In the following scenario, the difficulty is largely raised and the obstacle will have highly stochastic or even aggressive behaviors. In Fig. \ref{fig:sine curve} and Fig.  \ref{fig:sine_snap}, the obstacle passes in a sine curve, where the robot has an abrupt stop and right turn to avoid it. This result is actually quite sensible. In the beginning, the obstacle moves towards the positive $x$-axis and does not seem to collide with the robot soon. But the obstacle suddenly turns left at about (4, -1) and moves towards the robot.  Our robot thus stops quickly and turns right to avoid it. 

% Then, we let the obstacle deliberately block the robot. As shown in Fig. \ref{fig:obstruct} and Fig. \ref{fig:block_snap}, at first the robot passes around the coming obstacle. But the obstacle deliberately moves back at about (5, 0) and blocks the robot from returning to the reference path. The robot thus has to keep moving in the wrong direction and finally finds the opportunity to return at about (5, -1.5). 

% \begin{figure}[H]
% \centerline{\includegraphics[width=0.3\textwidth]{Obstacle passes in sine curve.pdf}}
% \caption{Obstacle passes in a sine curve.}
% \label{fig:sine curve}
% \end{figure}

% \begin{figure}[H]
% \centerline{\includegraphics[width=0.3\textwidth]{Obstacle blocks aggressively.pdf}}
% \caption{Obstacle blocks aggressively.}
% \label{fig:obstruct}
% \end{figure}

\section{Conclusion}
\label{sec:Conclusion}
We presented a model-based RL algorithm SPIL for chance-constrained policy optimization. Based on a feedback control view, we first reviewed and unified two existing chance-constrained RL methods to formulate a proportional-integral Lagrangian method, and enhanced it with an integral separation technique to prevent policy over-conservatism. To accelerate training, it also adopted a model-based gradient of safe probability for efficient policy optimization. We demonstrated the benefits of SPIL over previous methods in a car-following simulation. To prove its practicality, it was also applied to a real-world robot navigation task, where it successfully tracked the reference path while avoiding a highly stochastic moving obstacle.   
In the future, we will explore the possibility of online training, where the model and policy are both updated depending on the data from the real world to improve its online performance.

\bibliographystyle{ieeetr}
\bibliography{ref}

\end{document}